\documentclass{article} 
\usepackage{iclr2027_conference,times}

\usepackage{amsmath,amsfonts,bm}

\def\eqref#1{equation~\ref{#1}}

\def\1{\bm{1}}

\DeclareMathAlphabet{\mathsfit}{\encodingdefault}{\sfdefault}{m}{sl}
\SetMathAlphabet{\mathsfit}{bold}{\encodingdefault}{\sfdefault}{bx}{n}

\usepackage{hyperref}
\usepackage{url}
\usepackage{booktabs}
\usepackage{algorithmic}
\usepackage{booktabs}
\usepackage{graphicx}
\usepackage{booktabs}
\usepackage{tabularx}
\usepackage{array}
\usepackage{adjustbox}
\usepackage[utf8]{inputenc}
\usepackage[T1]{fontenc}
\usepackage{amsmath,amsthm,amssymb,mathtools}
\usepackage{amsfonts}
\usepackage{algorithm}
\usepackage{booktabs}
\usepackage{enumitem}
\usepackage{graphicx}
\usepackage{nicefrac}
\usepackage{microtype}
\usepackage{xcolor}
\usepackage{hyperref}
\usepackage{url}
\usepackage{booktabs}
\usepackage{bbm}

\makeatletter
\def\@oddhead{%
  \vbox{%
    \hbox to \textwidth{ \small \hfill  \hfill}%
    \vskip 4pt 
    \hrule height 0.5pt width \textwidth 
  }%
}
\def\@evenhead{\@oddhead}
\makeatother

\title{ORACLE: Optimizer-Relative Alignment for Constrained LEarning}

\author{
\textbf{Utkarsh Grover}\textsuperscript{1},
\textbf{Wyatt Mackey}\textsuperscript{2},
\textbf{Kaixun Hua}\textsuperscript{1},
\textbf{J. Morris Chang}\textsuperscript{1},
\textbf{Xiaomin Lin}\textsuperscript{1}
\\[0.6em]
\small
\textsuperscript{1}University of South Florida
\qquad
\textsuperscript{2}DEVCOM Army Research Lab
\\[0.4em]
\footnotesize
\texttt{\{utkarshgrover, khua, chang5, xlin2\}@usf.edu}
\\
\texttt{wyatt.t.mackey.civ@army.mil}
}

\iclrfinalcopy 
\begin{document}

\maketitle

\begin{abstract}
Constraint-handling methods typically intervene before the optimizer acts, by modifying the objective or the gradient. Yet momentum, adaptive scaling, and structured preconditioning can substantially reshape that signal before it becomes a parameter update. We formulate \emph{optimizer relative constrained learning}, where constraint compatibility is assessed on the post-optimizer update. Building on this view, we introduce \textsc{ORACLE}, which evaluates the native optimizer's realized step through a joint endpoint linearization of heterogeneous constraint families, constructs the resulting alignment in the optimizer's own geometry, bounds its authority, and commits it only after validation. We evaluate ORACLE across eight Partial Differential Equation benchmarks and four optimizers spanning Euclidean, diagonal adaptive, and structured preconditioned geometries, where it improves or matches native optimizer in 94\% of configurations. Cross-model analysis shows the same behavior in 92\% of configurations, while matched comparisons show improvements over alternative constraint-handling methods acting at the objective, gradient, and post-optimizer levels. 
\end{abstract}

\section{Introduction}
\vspace{-0.2cm}
Modern learning systems increasingly operate under requirements that extend beyond
minimizing a single task loss. Physics-informed learning must reconcile data with
governing equations and structural constraints~\cite{wang2026gradient,moro2025solving};
safety-critical robotics must optimize behavior while satisfying safety
requirements~\cite{nguyen2024gameplay,zhang2026safevla}; and decision-focused learning couples prediction with constrained downstream optimization, where predictive accuracy alone may not yield feasible or high-quality
decisions~\cite{mandi2024decision,mandi2026feasibility,kong2023df2}. In each case the learner must improve task performance while remaining compatible with additional physical, safety, or decision-level requirements.

The most direct way to impose such requirements is to write them into the quantity being minimized. Penalty terms, constraint multipliers, primal-dual formulations, and augmented-Lagrangian methods encode constraints in the objective itself
\cite{moro2025solving,nandwani2019primal,dener2020training,fioretto2020lagrangian}, and adaptive weighting schemes adjust their relative influence as training proceeds \cite{bischof2025multi}. These methods control how much each requirement contributes to a scalar trade-off, but they cannot express which \emph{directions} of parameter movement respect it: competing terms may remain directionally incompatible even after their magnitudes are balanced \cite{wang2026gradient}.

A second family therefore acts on directions, the gradient itself. Multi-objective
and conflict-resolution methods remove or reconcile components of competing gradients before a step is taken \cite{yu2020gradient,liu2021conflict,liu2025config,bu2026harmonized},
and projection or SQP-type methods correct a proposed movement using local constraint geometry \cite{rosen1961gradient,byrd2008inexact}. This resolves directional conflict that reweighting cannot. Both families, however, hand their corrected signal to an optimizer, and treat that handover as though the optimizer will apply what it receives.

Each optimizer defines its own geometry by scaling and prioritizing parameter magnitude and directions differently. Momentum, adaptive coordinate statistics, and structured preconditioning transform the incoming gradient using information accumulated across training, so the displacement ultimately applied to the parameters can differ from that gradient in both scale and direction
\cite{kingma2014adam,gupta2018shampoo,vyas2025soap,si2025adamuon,du2026newton}. Adaptive optimizers descend in a geometry that is not Euclidean \cite{xie2026tale}, and recent work names the resulting gradient-update mismatch  directly in physics-informed training \cite{xiao2026gradient}. A constraint enforced before this transformation is therefore enforced in a geometry that is not the geometry of the step actually taken, and this gap widens precisely for the stateful, preconditioned optimizers on which modern training relies. This leaves one position in the pipeline unoccupied: the realized update, after the optimizer has committed to how the parameters should move.

\begin{figure*}[t]
\centering
\includegraphics[width=\textwidth]{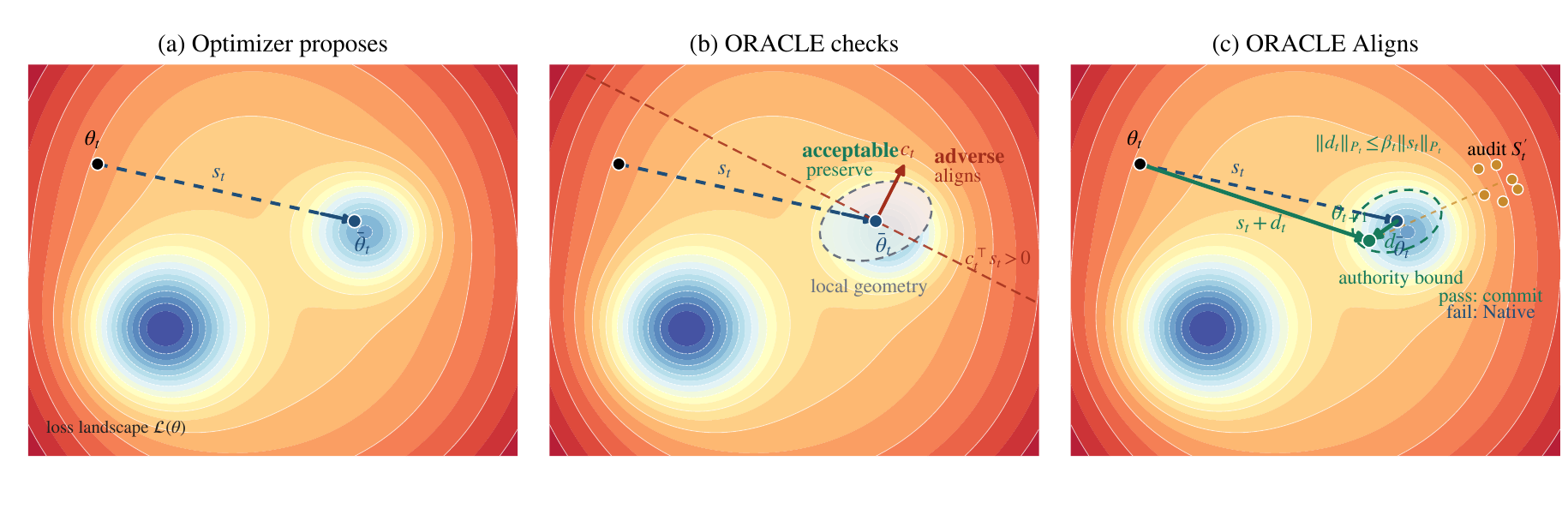}
\vspace{-0.7cm}
\caption{\textbf{Toy illustration of ORACLE.}
(a) From $\theta_t$, the native optimizer proposes the endpoint
$\bar{\theta}_t$.
(b) ORACLE checks the proposal against the local constraint geometry.
(c) A validated alignment produces $\theta_{t+1}$; otherwise the native
proposal is retained.}
\label{fig:intro}
\vspace{-0.4cm}
\end{figure*}

We introduce \textsc{ORACLE} (\emph{Optimizer-Relative Alignment for Constrained
LEarning}), illustrated in Figure~\ref{fig:intro}, which occupies that position. From $\theta_t$, the native optimizer uses its internal state to produce the realized step $s_t$ and proposed endpoint $\bar{\theta}_t=\theta_t+s_t$. At this endpoint,
ORACLE jointly linearizes the sampled constraint families so that their local responses to the same parameter movement are considered together. The resulting local model is then used to construct an alignment in the geometry of the optimizer that produced $s_t$. The quantity $c_t^\top s_t$ identifies locally adverse realized movements; when it is positive, ORACLE constructs a bounded optimizer-relative alignment and commits it only after nonlinear validation and an independent audit. Thus, $\theta_{t+1}=\theta_t+s_t+d_t$ when the alignment is accepted and $\theta_{t+1}=\bar{\theta}_t$ otherwise.

We study this in physics-informed Partial Differential Equation learning, where the gap is most exposed.
PINNs~\cite{raissi2019physics} combine PDE residuals with initial and boundary conditions
whose gradients differ sharply in scale, conditioning, and direction, and modern PINN
training increasingly depends on adaptive and structured
optimizers~\cite{wang2026gradient,rathore2024challenges}. Heterogeneous constraint
sensitivities therefore meet stateful preconditioned updates in the same training loop. We
ask five questions of this setting: whether optimizer-relative alignment holds across
native optimizer geometries; whether it persists when the model formulation changes; what's
the additional computation costs and what it buys; whether intervening on the realized
update outperforms intervening earlier under a matched protocol; and which of ORACLE's
mechanisms are responsible.

Our contributions are as follows. 
\textbf{(i)} We formulate \emph{optimizer-relative constrained learning}, in which constraint compatibility is defined on the realized update rather than the gradient preceding it. \textbf{(ii)} We instantiate it as ORACLE, which jointly linearizes heterogeneous
constraint families at the proposed endpoint and constructs a single alignment in the certified geometry of the native optimizer. A matrix-free residual-space
solve computes this alignment without forming the parameter-space Jacobian, while native-relative authority and held-out validation control whether it is committed. \textbf{(iii)} We evaluate it across optimizer geometries, model formulations, computational cost, intervention points, and internal mechanisms, improving or matching the native optimizer in 94\% valid PDE optimizer configurations and 92\% model optimizer configurations.

\section{Related Work}
\vspace{-0.2cm}
Modern optimizers separate the raw gradient from the parameter displacement ultimately applied to the model. AdaGrad, RMSProp, Adam, and Adafactor use gradient history for coordinate-wise adaptation \cite{ward2020adagrad,kingma2014adam,shazeer2018adafactor}, while natural gradient, K-FAC, Shampoo, SOAP, and Newton-type methods introduce metric, structured-preconditioning, or curvature information \cite{kakade2001natural,grosse2016kronecker,gupta2018shampoo,vyas2025soap,botev2017practical}. The realized update can therefore differ from the raw gradient in both direction and scale, making the optimizer's induced geometry consequential for where and how constraint information is
introduced.

Penalty, Lagrangian, primal-dual, and augmented-Lagrangian methods encode
constraints directly in the optimization problem
\cite{nandwani2019primal,dener2020training,lavado2023achieving,fioretto2020lagrangian},
while multi-objective methods reconcile competing gradients
\cite{yu2020gradient,liu2021conflict,navon2022multi} and
projection or SQP-type methods correct candidate movement using local
constraint geometry \cite{rosen1961gradient,byrd2008inexact}.
These approaches span objective,
gradient, and movement-level intervention, but generally do not make the native optimizer's realized geometry the reference geometry for constraint correction.

Recent methods enforce nonlinear requirements through constrained formulations and feasibility mechanisms
~\cite{moro2025solving,lastrucci2025enforce,iftakher2025physics}, or by
modifying task gradients~\cite{liu2025config,bu2026harmonized}. Yet in PINNs, heterogeneous constraint sensitivities already induce difficult
optimization geometry~\cite{wang2021understanding,yao2023multiadam,
rathore2024challenges,wang2026gradient}, which adaptive and structured
optimizers can further reshape~\cite{kingma2014adam,vyas2025soap}.
This creates a missing interface between nonlinear
constraint geometry and the geometry of the realized optimizer update, together with the problem of
deciding when a locally constructed correction should be trusted.

ORACLE addresses this interface by recovering and certifying the native optimizer’s realized metric
and expressing sampled residual geometry within it. It applies a bounded correction only when the
realized step is locally adverse, then separates candidate construction from acceptance through an
independent audit support. Detailed comparisons with post-optimizer
alignment, projection methods, and sample-splitting approaches appear in the Extended Related Work ~\ref{app:related_work}.

\vspace{-0.2cm}
\section{ORACLE: Optimizer-Relative Alignment for Constrained LEarning}
\label{sec:method}
\vspace{-0.2cm}

ORACLE wraps a native optimizer. At iteration $t$ the optimizer proposes
moving from $\theta_t$ to $\bar\theta_t=\theta_t+s_t$; ORACLE commits either that proposal or a corrected one,
\begin{equation}
    \boxed{\;
    \theta_{t+1}=\theta_t+s_t+d_t,
    \qquad
    d_t=\mathbf 1\{\mathrm{Cert}_t\}\,
        \mathbf 1\{\mathrm{Trig}_t\}\,
        \mathbf 1\{\mathrm{Val}_t\}\;
        \alpha_t\,\widehat d_t .
    \;}
    \label{eq:oracle_update}
\end{equation}
Each factor of $d_t$ is one mechanism (Fig.~\ref{fig:intro}). The
\emph{alignment direction} $\widehat d_t$ reduces the local constraint model
in the native optimizer's own geometry; the \emph{corrective authority}
$\alpha_t$ bounds how far the proposal may be redirected; and three gates
decide whether the correction is applied: \emph{geometry certification}
$\mathrm{Cert}_t$, the \emph{alignment trigger} $\mathrm{Trig}_t$, and
\emph{validation} $\mathrm{Val}_t$. If any gate is closed, $d_t=0$ and ORACLE
returns the native step exactly. Algorithm~\ref{alg:oracle} lists the
procedure; the rest of this section defines each factor in turn.

Constraints are organized into $K$ semantic families. In a PINN, for example,
these may correspond to the PDE residual, boundary conditions, initial
conditions, or observed data. For family $k$, $r_k(\theta;z)$ measures the
violation at evaluation point $z$, $z\sim\nu_k$ specifies how those points
are sampled, and $\omega_{k,t}>0$ controls the family's relative
contribution.

\vspace{-0.1cm}
\subsection{Native Proposal and Constraint Model}
\label{sec:native_proposal}
\vspace{-0.1cm}

Let $g_t$ be the gradient supplied to the native optimizer, and $\xi_t$ its
internal state, such as momentum, running moments, or structured
preconditioning statistics. The native optimizer produces
\begin{equation}
    s_t=\mathcal S(g_t,\xi_t),
    \qquad
    \xi_{t+1}=\mathcal U(g_t,\xi_t),
    \qquad
    \bar\theta_t=\theta_t+s_t,
    \label{eq:native_optimizer}
\end{equation}
where $s_t$ is the realized parameter displacement and $\bar\theta_t$ the
native proposal. ORACLE leaves both $s_t$ and the state transition untouched.
With $d_t=0$, Eq.~\ref{eq:oracle_update} is the native update
$\theta_{t+1}=\bar\theta_t$; this is the default that each mechanism below
modifies only under a stated condition.

To evaluate this proposal, we summarize the population-level constraint violation as
\begin{equation}
    \Phi_t(\theta)
    =
    \sum_{k=1}^{K}\omega_{k,t}\,
    \mathbb E_{z\sim\nu_k}\!\left[r_k(\theta;z)^2\right].
    \label{eq:population_constraint_energy}
\end{equation}
After $\bar\theta_t$ is fixed, ORACLE evaluates its constraint violations on
a fresh support: for each family $k$ it draws $m_k$ samples
$z_{k,j}\sim\nu_k$, collects them in $S_t$, stacks the residuals into
$r_{S_t}(\theta)$, and measures
\begin{equation}
    \Phi_{S_t}(\theta)
    =
    \tfrac12\,r_{S_t}(\theta)^\top W_t\,r_{S_t}(\theta),
    \label{eq:sampled_constraint_energy}
\end{equation}
where $W_t$ is diagonal and assigns each residual from family $k$ the weight
$2\omega_{k,t}/m_k$. Because $\bar\theta_t$ is fixed before $S_t$ is sampled,
$\Phi_{S_t}(\bar\theta_t)$ is an unbiased estimate of the population
constraint energy (Appendix~\ref{app:sampling_unbiasedness}). The sampled energy tells us how much violation remains at the proposed endpoint. Since $r_{S_t}$ stacks PDE, boundary, initial-condition, and other residuals into a single endpoint model, ORACLE next asks how these constraint families
respond jointly to the same parameter movement. Answering this requires their local sensitivities and the geometry in which the optimizer produced $s_t$.

\vspace{-0.1cm}
\subsection{Checking the Proposal in the Native Optimizer Geometry}
\label{sec:geometry_trigger}
\vspace{-0.1cm}

Different optimizers induce different geometries on parameter updates: Adam
rescales coordinates using adaptive statistics, while SOAP and PSGD-Kron
additionally apply structured preconditioning. Rather than constructing a
new geometry, ORACLE reuses the transformation already encoded by the native
optimizer and writes the realized step as
\begin{equation}
    s_t=-B_tu_t,
    \qquad
    P_t=B_t^{-1}\succ0,
    \label{eq:certified_metric}
\end{equation}
where $u_t$ is the optimizer's effective direction and $B_t$ is the symmetric
positive-definite (SPD) scaling or preconditioning operator induced by its
update rule and state. For SGD, $B_t$ is scalar learning-rate scaling; for
Adam, diagonal adaptive scaling; for structured optimizers, it acts through
their existing factors. The inverse $P_t$ defines the optimizer-relative
metric $\|v\|_{P_t}^2=v^\top P_tv$, which reflects the native optimizer's
directional preferences. Neither $B_t$ nor $P_t$ is formed explicitly;
ORACLE uses them only through optimizer-specific vector products. The
geometry is meaningful only if it faithfully describes the step actually
produced, so ORACLE certifies it against the realized displacement,
\begin{equation}
    \|s_t+B_tu_t\|_2
    \le
    \tau_{\mathrm{rel}}\|s_t\|_2+\tau_{\mathrm{abs}},
    \label{eq:metric_certification}
\end{equation}
together with optimizer-specific validity checks
(Appendix~\ref{app:optimizer_adapters}); $\mathrm{Cert}_t$ is the event that
both hold. On failure ORACLE preserves the native proposal and never replaces
an uncertified geometry with $P_t=I$. This supplies the gate
$\mathbf 1\{\mathrm{Cert}_t\}$ and the norm in which every quantity below is
measured.

ORACLE now asks whether the movement from $\theta_t$ to $\bar\theta_t$ is
locally adverse to the sampled constraints. At the proposed endpoint, define
\begin{equation}
    r_t=r_{S_t}(\bar\theta_t),
    \qquad
    J_t=\left.\frac{\partial r_{S_t}(\theta)}{\partial\theta}
        \right|_{\theta=\bar\theta_t},
    \qquad
    c_t=J_t^\top W_tr_t=\nabla\Phi_{S_t}(\bar\theta_t),
    \label{eq:endpoint_geometry}
\end{equation}
where $r_t$ measures the sampled violations at the endpoint, $J_t$ their
local sensitivity to parameter movement, and $c_t$ the first-order direction
of increasing constraint energy. The trigger is the directional interaction
between $c_t$ and the realized step:
\begin{equation}
    \mathrm{Trig}_t=\{\,c_t^\top s_t>0\,\}.
    \label{eq:oracle_trigger}
\end{equation}
When it fires, the native movement is locally adverse to the sampled
constraint model and ORACLE opens the alignment path; otherwise the native
proposal is preserved. $J_t$ is never formed explicitly: a Jacobian-vector
product $J_tv$ measures how a parameter movement $v$ changes the residuals,
and a vector-Jacobian product $J_t^\top q$ maps a residual-space direction
back to parameter space, both by automatic differentiation. Formal analysis
of the endpoint trigger appears in Appendix~\ref{app:endpoint_signal}. This
supplies the gate $\mathbf 1\{\mathrm{Trig}_t\}$.

\vspace{-0.1cm}
\subsection{Optimizer-Relative Alignment and Validated Commitment}
\label{sec:alignment_commit}
\vspace{-0.1cm}

Once the trigger fires, the certified metric determines how the proposal is
corrected. At $\bar\theta_t$, let $b_t=W_t^{1/2}r_t$ and
$C_t=W_t^{1/2}J_t$. ORACLE computes the alignment direction by solving
\begin{equation}
    \widehat d_t
    =
    \arg\min_{d\in\mathbb R^p}\
    \tfrac12\|b_t+C_td\|_2^2
    +\tfrac{\lambda_t}{2}\|d\|_{P_t}^2 ,
    \label{eq:alignment_primal}
\end{equation}
with damping $\lambda_t>0$. The first term jointly reduces the local
linearized violations of the stacked constraint families; the second keeps the adjustment close to the native proposal in the optimizer's own geometry. Rather than solving Eq.~\ref{eq:alignment_primal} in parameter space, ORACLE uses
\begin{equation}
    G_t=C_tP_t^{-1}C_t^\top
    =W_t^{1/2}J_tP_t^{-1}J_t^\top W_t^{1/2},
    \qquad
    (G_t+\lambda_tI)\,z_t=b_t,
    \qquad
    \widehat d_t=-P_t^{-1}C_t^\top z_t .
    \label{eq:oracle_alignment}
\end{equation}
Neither $J_t$ nor $G_t$ is materialized: a product with $G_t$ uses one VJP, one application of $P_t^{-1}$, and one JVP. The resulting direction is a strict first-order constraint-descent direction;
Appendix~\ref{app:alignment_derivation} gives the exact and finite-CG
results. This supplies the direction of $d_t$.

ORACLE next limits how far the constraints may redirect the native proposal,
scaling $\widehat d_t$ by
\begin{equation}
    \alpha_t
    =
    \min\!\left\{
        \frac{c_t^\top s_t}{-c_t^\top\widehat d_t+\varepsilon_{\mathrm{auth}}},
        \;
        \frac{\beta_t\|s_t\|_{P_t}}{\|\widehat d_t\|_{P_t}}
    \right\},
    \qquad
    \widetilde d_t=\alpha_t\widehat d_t ,
    \label{eq:authority}
\end{equation}
where $\beta_t\in[0,1)$ controls the maximum corrective authority relative to
the native step and $\varepsilon_{\mathrm{auth}}>0$ is a small numerical
stabilizer. The first term is the scale needed to neutralize the detected
first-order adverse signal; the second caps the correction in the
optimizer-relative metric. Consequently
$\|\widetilde d_t\|_{P_t}\le\beta_t\|s_t\|_{P_t}$, so the aligned update
remains bounded relative to the movement selected by the native optimizer.
This supplies the magnitude of $d_t$: $d_t\in\{0,\widetilde d_t\}$.

Because the alignment is derived from a local linear model, ORACLE verifies
that its predicted improvement persists under the nonlinear constraints,
\begin{equation}
    \mathrm{Pred}_t
    =
    \Phi_{S_t}(\bar{\theta}_t)
    -\frac12\lVert b_t+C_t\widetilde d_t\rVert_2^2,
    \qquad
    \mathrm{Ared}_t
    =
    \Phi_{S_t}(\bar{\theta}_t)
    -\Phi_{S_t}(\bar{\theta}_t+\widetilde d_t).
    \label{eq:pred_ared}
\end{equation}
and the candidate proceeds only if $\mathrm{Pred}_t>0$, $\mathrm{Ared}_t>0$,
and $\mathrm{Ared}_t/\mathrm{Pred}_t\ge\rho_{\min}$. Because this check
reuses $S_t$, it tests local-model agreement rather than independent
improvement. ORACLE therefore fixes the candidate and evaluates it on an
independent audit support $S_t'$, accepting only if
\begin{equation}
    \Phi_{S_t'}(\bar\theta_t+\widetilde d_t)
    \le
    \Phi_{S_t'}(\bar\theta_t).
    \label{eq:audit_gate}
\end{equation}
$\mathrm{Val}_t$ is the event that both checks pass. Finally,
\begin{equation}
    d_t=
    \begin{cases}
        \widetilde d_t, & \text{if all checks pass},\\
        0,              & \text{otherwise},
    \end{cases}
    \qquad
    \theta_{t+1}=\theta_t+s_t+d_t ,
    \label{eq:oracle_commit}
\end{equation}
which is~\eqref{eq:oracle_update}. Any rejection recovers the exact native update while leaving $\xi_{t+1}$ unchanged. In summary, the native optimizer \emph{proposes}, the constraints \emph{check}, ORACLE \emph{aligns} in the certified native geometry, and validation determines whether the aligned proposal is committed.

\begin{algorithm}[t]
\caption{\textsc{ORACLE}: Optimizer-Relative Alignment for Constrained Learning}
\label{alg:oracle}
\begin{algorithmic}[1]
\REQUIRE Parameters $\theta_t$, optimizer state $\xi_t$, task gradient $g_t$;
\REQUIRE constraint sampling laws and ORACLE hyperparameters
\STATE Form the native proposal
$(s_t,\xi_{t+1}) \leftarrow \mathrm{NativeStep}(g_t,\xi_t)$,
$\bar{\theta}_t \leftarrow \theta_t+s_t$
\STATE Recover and certify the optimizer geometry $(B_t,P_t,u_t)$
via Eq.~\ref{eq:certified_metric}- Eq.~\ref{eq:metric_certification};
\textbf{return} $(\bar{\theta}_t,\xi_{t+1})$ if certification fails
\STATE Draw construction support $S_t$ and evaluate the local constraint model
at $\bar{\theta}_t$ to obtain $(r_t,c_t)$ and Jacobian products with $J_t$
\STATE \textbf{return} $(\bar{\theta}_t,\xi_{t+1})$
if $c_t^\top s_t\le 0$ \COMMENT{trigger Eq.~\ref{eq:oracle_trigger}}
\STATE Construct the optimizer-relative alignment direction
$\widehat d_t$ using Eq.~\ref{eq:oracle_alignment}
\STATE \textbf{return} $(\bar{\theta}_t,\xi_{t+1})$
if $c_t^\top\widehat d_t\ge 0$
\COMMENT{finite-precision guard}
\STATE Apply the native-relative authority bound
Eq.~\ref{eq:authority} to obtain $\widetilde d_t$
\STATE Validate $\widetilde d_t$ on $S_t$ using
Eq.~\ref{eq:pred_ared};
\textbf{return} $(\bar{\theta}_t,\xi_{t+1})$ if validation fails
\STATE Fix $\widetilde d_t$, draw independent audit support $S'_t$, and
\textbf{return} $(\bar{\theta}_t,\xi_{t+1})$ if the held-out audit
Eq.~\ref{eq:audit_gate} fails
\STATE \textbf{return}
$(\bar{\theta}_t+\widetilde d_t,\xi_{t+1})$
\end{algorithmic}
\end{algorithm}

\vspace{-0.2cm}
\section{Experiment}
\label{sec:experiments}
\vspace{-0.2cm}
The central claim of ORACLE is that constraint alignment should be defined relative to the movement produced by the native optimizer. We test this claim from breadth to mechanism, and structure our experiments around the following research questions: Q1 varies optimizer geometry; Q2 varies model formulation; Q3 quantifies the cost and solution-level benefit of alignment; Q4 compares objective, gradient, and post-optimizer intervention; and Q5 isolates ORACLE's geometry, authority, and validation mechanisms.

Within every Native ORACLE pair, we hold the architecture, native objective,
optimizer configuration, learning-rate schedule, sampling pipeline, and
training budget fixed; ORACLE acts only after the native optimizer forms its
realized update. Our primary metric is relative $\ell_2$ solution error
(rel-L2), with lower values indicating greater solution accuracy. Complete
architectures, optimizer settings, sampling protocols, and ORACLE
hyperparameters appear in Appendix~\ref{app:experimental_details}.

\begin{figure*}[t]
\centering
\includegraphics[width=\textwidth]{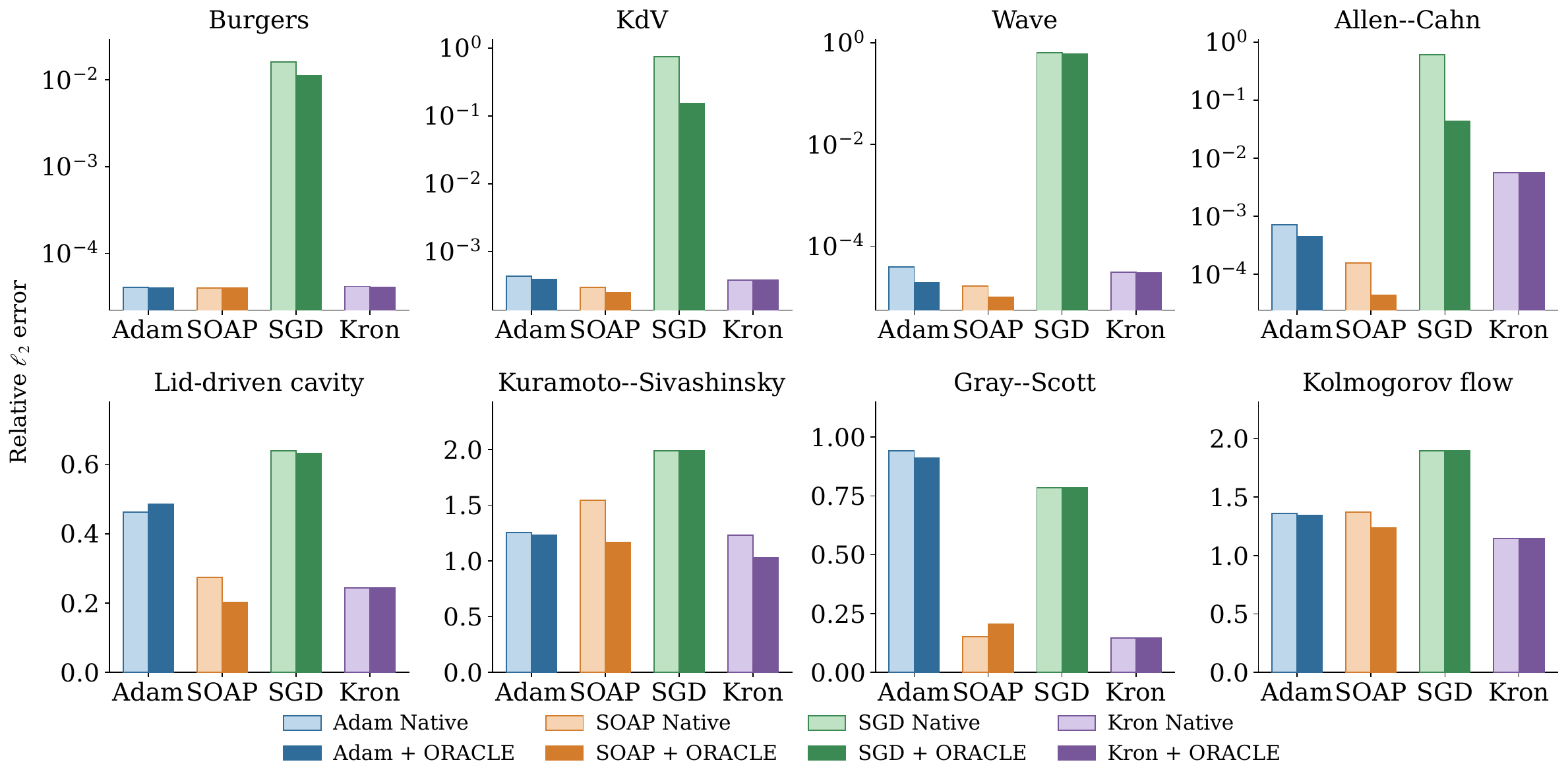}
\vspace{-0.6cm}
\caption{
\textbf{Performance across optimizer geometries.}
Endpoint rel-$L_2$ error across eight PDE benchmarks for Adam, SOAP, SGD, and PSGD-Kron, comparing each Native VS Native+ORACLE. Bars show means; the standard deviation across three independent runs is $<5\%$ for all reported entries.
}
\label{fig:optimizer_generality}
\vspace{-0.4cm}
\end{figure*}

\vspace{-0.1cm}
\subsection {Q1: Does ORACLE Work Across Optimizer Geometries?}
\vspace{-0.1cm}
ORACLE defines alignment relative to the geometry of the native optimizer.
Its first test is therefore whether the same alignment principle remains effective when that geometry changes. We pair ORACLE with
SGD~\cite{robbins1951stochastic}, Adam~\cite{kingma2014adam},
SOAP~\cite{vyas2025soap}, and PSGD-Kron~\cite{li2017preconditioned}
across eight PDE benchmarks, fixing PirateNet~\cite{wang2024piratenets}
throughout. These optimizers span Euclidean, diagonal-adaptive, and structured
preconditioned update geometries.

\vspace{-0.1cm}
\paragraph{Results.}
\vspace{-0.1cm}
Figure~\ref{fig:optimizer_generality} shows that ORACLE improves or matches
Native in $30/32$ valid PDE optimizer pairs: $23$ improve, $7$ match, and
$2$ degrade. Improvements occur under every optimizer family, spanning
Euclidean SGD, diagonal-adaptive Adam, and the structured geometries of SOAP
and PSGD-Kron. ORACLE therefore does not depend on a single optimizer or a
shared update geometry.

Q1 establishes breadth across native optimizer geometries, but all runs use
PirateNet. Q2 tests whether the same behavior persists when the model
formulation changes. Complete results appear in
Appendix~\ref{app:q1_optimizer_geometry}.

\begin{figure*}[t]
\centering
\includegraphics[width=\textwidth]{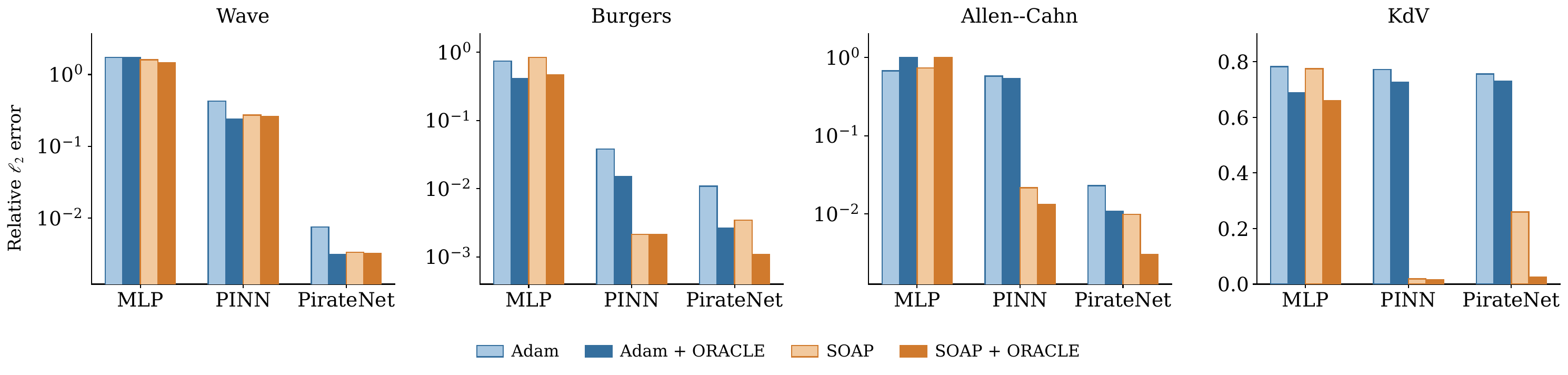}
\vspace{-0.6cm}
\caption{
\textbf{Performance across model formulations.}
Endpoint rel-$L_2$ error on Wave, Burgers, Allen-Cahn, and KdV for MLP, PINN, and PirateNet under Adam and SOAP, VS ORACLE. Lower is better. Each entry is mean, the std-dev across three independent runs < 7\% for reported entries.
}
\label{fig:model_generality}
\vspace{-0.2cm}
\end{figure*}

\vspace{-0.1cm}
\subsection{Q2: Does Alignment Persist Across Model Formulations?}
\vspace{-0.1cm}
Q1 varies the optimizer while fixing PirateNet, leaving open whether the
effect depends on that model. We therefore evaluate a standard MLP, a
conventional PINN~\cite{raissi2019physics}, and
PirateNet~\cite{wang2024piratenets} on Wave, Burgers, Allen-Cahn, and KdV
under Adam and SOAP. The MLP and PINN share the same backbone but differ in
whether the PDE residual enters the native objective, while PirateNet changes
both the architecture and its physics-specific inductive structure.

\vspace{-0.1cm}
\paragraph{Results.}
\vspace{-0.1cm}
Figure~\ref{fig:model_generality} shows that ORACLE improves or matches
Native in $22/24$ configurations. The effect spans all three model
formulations and both native optimizers, with gains reaching $60.6\%$ for
PINN+Adam on Burgers and $90.6\%$ for PirateNet+SOAP on KdV.

The two degradations occur for Pure MLP on Allen-Cahn under Adam and SOAP,
showing that ORACLE does not override limitations of the underlying learning
formulation. Nevertheless, Q2 extends the Q1 result beyond PirateNet and
shows that optimizer-relative alignment persists across substantially
different model formulations.

\begin{figure*}[t]
\centering
\includegraphics[width=\textwidth]{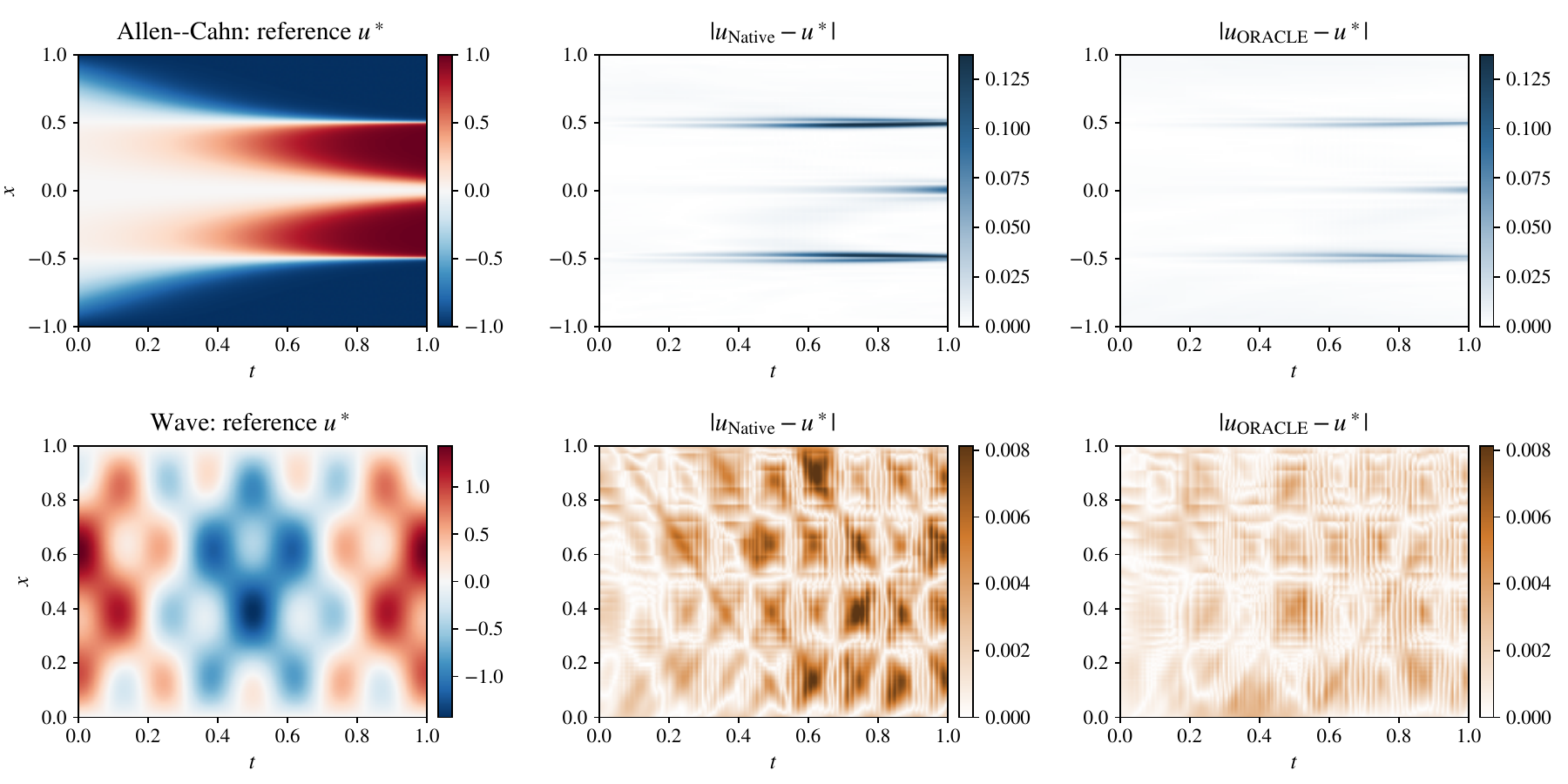}
\vspace{-0.6cm}
\caption{
\textbf{Representative solution and error fields.}
Reference solution (left), Native pointwise error (middle), and ORACLE
pointwise error (right) for Allen--Cahn and Wave. ORACLE reduces
interface localized error on Allen--Cahn and broader space time error on
Wave. Native and ORACLE use the same error scale within each PDE. 
}
\label{fig:error_field}
\vspace{-0.2cm}
\end{figure*}

\vspace{-0.1cm}
\paragraph{Where do the gains appear?}
\vspace{-0.1cm}
Aggregate rel-$L_2$ obscures where solution error is reduced. Figure~\ref{fig:error_field} compares the reference solution with matched
Native and ORACLE pointwise error fields. On Allen--Cahn, Native error is
concentrated along the phase interfaces, where ORACLE substantially attenuates
the error bands. On Wave, the reduction is distributed more broadly across
space time. Thus, the endpoint gains reflect meaningful changes in the learned
solution rather than only improvements in an aggregate metric. Complete results appear in Appendix~\ref{app:q2_model_formulation}.


\begin{figure*}[t]
\centering
\includegraphics[width=\textwidth]{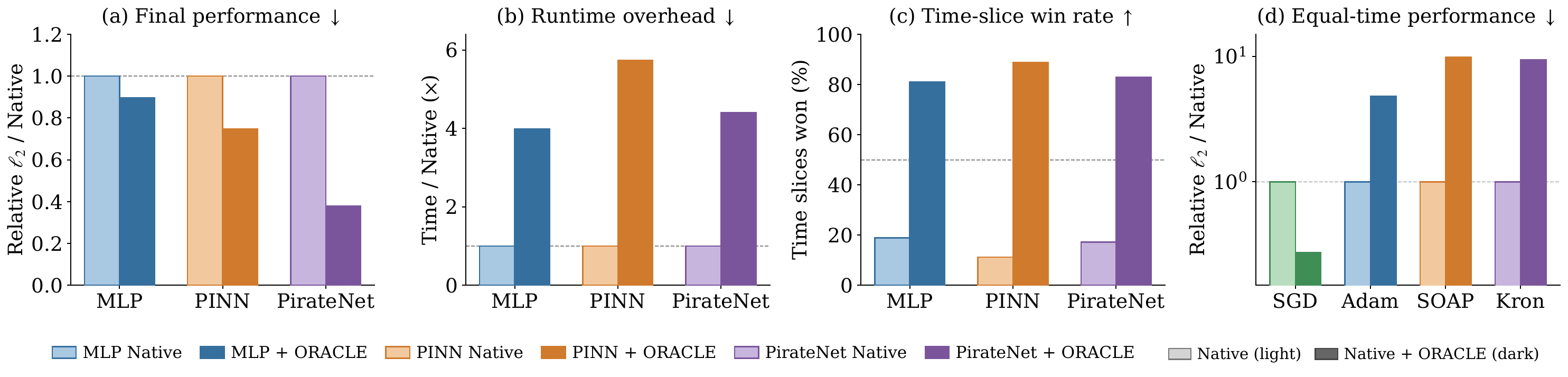}
\vspace{-0.6cm}
\caption{
\textbf{Cost-benefit of ORACLE.}
Endpoint accuracy, per-step runtime, temporal win rate, and equal-wall-clock
performance across model formulations and optimizer geometries. 
}
\label{fig:cost_benefit}
\vspace{-0.2cm}
\end{figure*}

\vspace{-0.1cm}
\subsection{Q3: What Does ORACLE Cost, and What Does That Cost Buy?}
\vspace{-0.1cm}
\label{sec:q3_cost}

Q1 and Q2 establish that ORACLE remains effective across optimizer geometries
and model formulations. We next ask what these gains cost, whether they persist
beyond endpoint error, and whether they amortize under a fixed compute budget.

\vspace{-0.1cm}
\paragraph{Results.}

Figure~\ref{fig:cost_benefit}(a) shows that ORACLE reduces normalized endpoint error to $0.90\times$, $0.75\times$, and $0.38\times$ Native for MLP, PINN, and PirateNet, respectively. These gains come at a cost: ORACLE requires approximately $4.0$--$5.8\times$ more time per optimization step.(Fig.~\ref{fig:cost_benefit}b).

Panel~\ref{fig:cost_benefit}(c) asks whether the improvement persists through the physical evolution of the PDE rather than appearing only in the final aggregate error. At each physical time slice, we compare Native and ORACLE using the relative
$\ell_2$ error over the complete spatial field; ORACLE wins a slice only when its full spatial prediction is more accurate. Across matched configurations, ORACLE wins $81.1\%$, $88.8\%$, and $82.9\%$ of evaluated time slices for
MLP, PINN, and PirateNet, respectively. Thus, the endpoint gains reflect persistent improvements throughout the predicted dynamics rather than a small set of favorable spatial locations or times.

Panel~\ref{fig:cost_benefit}(d) imposes the stricter equal-wall-clock test, giving Native and ORACLE the same elapsed time rather than the same number of optimization steps. This comparison combines update quality with update throughput. ORACLE still reduces aggregate error by approximately $73\%$ with SGD, showing that its alignment benefit can exceed its correction cost. For Adam, SOAP, and PSGD-Kron, the additional correction does not yet recover its runtime overhead within the same budget. 

This sharpens Q1: optimizer-relative alignment remains effective across all tested geometries, while its compute-normalized benefit depends on how quickly that geometry-aware correction can be realized. Together, the results separate \emph{alignment effectiveness} from \emph{compute efficiency}.
For detailed analysis see Appendix~\ref{app:cost_benefit}.

\vspace{-0.1cm}
\subsection{Q4: Where Should Constraint Intervention Occur?}
\vspace{-0.1cm}
Q1, Q2 and Q3 establish breadth, but do not determine whether the gains arise
from optimizer-relative alignment or merely from additional constraint
handling. We therefore compare objective, gradient, and post-optimizer
interventions on Allen-Cahn and Wave. Every method branches from the same
$5{,}000$-step Native and receives an identical additional
$5{,}000$ optimization steps, controlling initialization, optimizer state,
and training horizon.

\vspace{-0.1cm}
\paragraph{Results.}

\begin{figure*}[t]
\centering
\includegraphics[width=\textwidth]{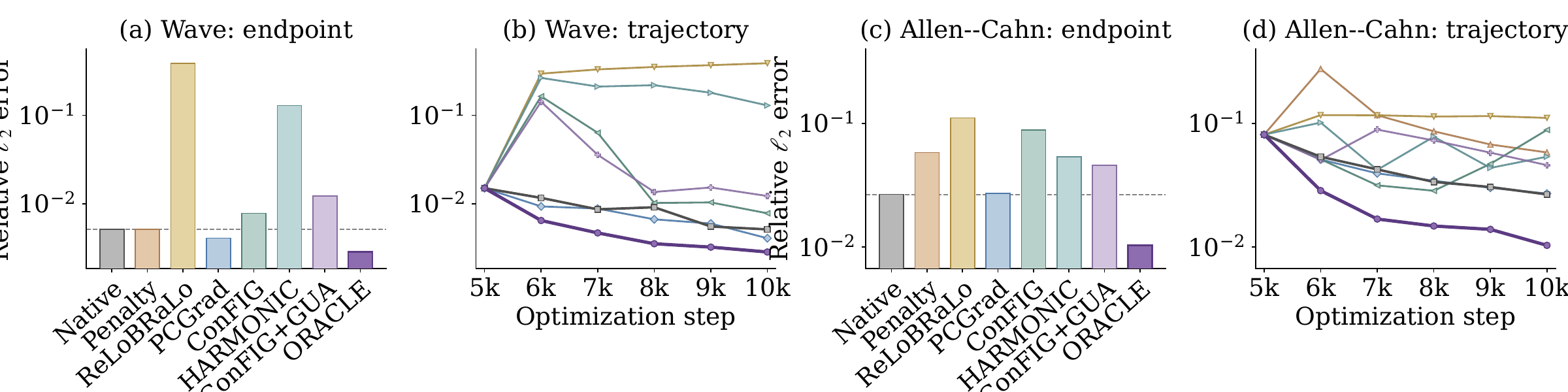}
\vspace{-0.6cm}
\caption{
\textbf{Matched Q4 intervention comparison on Wave and Allen-Cahn.}
Panels (a,c) show endpoint rel-$L_2$ error; panels (b,d) show trajectories from the shared $5$k Native checkpoint to $10$k. ORACLE achieves
the lowest endpoint error on both PDEs. Each entry is mean, the std-dev across three independent runs < 7\% for all reported entries.
}
\label{fig:q3_intervention}
\vspace{-0.2cm}
\end{figure*}

Figure~\ref{fig:q3_intervention} shows that ORACLE achieves the lowest endpoint rel-$L_2$ error on both PDEs, improving over Native by $61.2\%$ on Allen--Cahn and $44.6\%$ on Wave. The post-fork trajectories strengthen this result: from the shared $5$k Native, ORACLE remains below Native at every recorded milestone and decreases monotonically to the $10$k endpoint on both problems. Objective- and gradient-level interventions do not show the same consistency. Post-optimizer intervention alone does not explain which part of ORACLE produces the gain. Q5 therefore fixes \emph{where} constraints act and examines \emph{how} the resulting update is aligned.

\subsection{Q5: What Makes Optimizer-Relative Alignment Work?}
\label{sec:geometric_analysis}
\begin{figure*}[t]
\centering
\includegraphics[width=\textwidth]{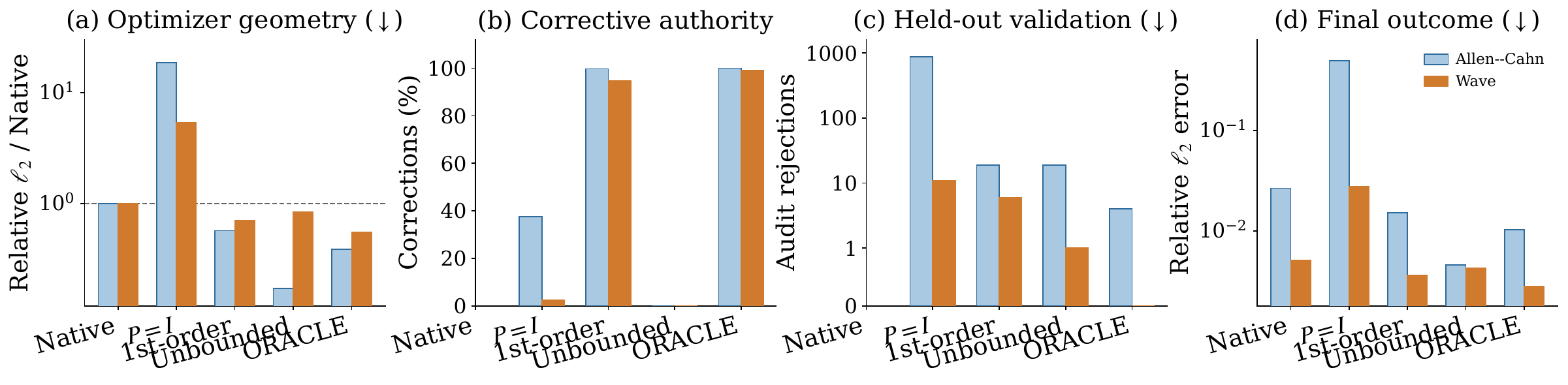}
\vspace{-0.4cm}
\caption{\textbf{Mechanism analysis of ORACLE on Allen--Cahn and Wave.}
(a) Native optimizer geometry is essential; (b) the authority bound is active for nearly all accepted alignments; (c) held-out validation rejects unreliable candidates; and (d) full ORACLE improves both PDEs more consistently}
\label{fig:geometric_analysis}
\vspace{-0.4cm}
\end{figure*}

\vspace{-0.1cm}
Q5 therefore examines three mechanisms: optimizer-relative geometry, corrective authority, and validation. We compare five settings on Allen-Cahn and Wave: \emph{Native} applies no post-optimizer correction; $P_t=I$ replaces the optimizer metric with Euclidean geometry; \emph{First-order} retains the native metric but uses the direct correction $d\propto-P_t^{-1}c_t$; \emph{Unbounded} removes the authority cap; and \emph{ORACLE} combines the full optimizer-relative, bounded, and validated alignment. Figure~\ref{fig:geometric_analysis} traces these choices from
alignment behavior to final solution error.

To compare how each mechanism affects the final solution on a common scale, we normalize endpoint error by the matched Native optimizer,
$
R_{\mathrm{err}}
=
\frac{\mathrm{relL2}(\mathrm{method})}
     {\mathrm{relL2}(\mathrm{Native})},
$
where $R_{\mathrm{err}}<1$ denotes improvement. Replacing the
optimizer-relative metric with $P_t=I$ increases error to
$18.6\times$ Native on Allen--Cahn and $5.4\times$ on Wave
(Fig.~\ref{fig:geometric_analysis}a). In contrast, the metric-aware
first-order variant and full ORACLE remain below Native on both PDEs.
Thus, successful alignment depends strongly on measuring movement in the
geometry induced by the native optimizer rather than in a common Euclidean
metric.
ORACLE bounds the aligned adjustment in the same geometry,
$
\|\widetilde d_t\|_{P_t}
\le
\beta_t\|s_t\|_{P_t}.
$
This bound is active on $100\%/99\%$ of accepted ORACLE alignments on
Allen-Cahn/Wave and $100\%/95\%$ for the metric-aware first-order variant,
compared with $38\%/2.5\%$ under $P_t=I$
(Fig.~\ref{fig:geometric_analysis}b). The optimizer metric therefore affects
not only the alignment direction but also what constitutes a permissible
adjustment relative to the native step.
ORACLE evaluates each fixed candidate on independent support before
commitment. Over $5$k active steps, the audit rejects $875$ candidates under $P_t=I$ versus only $4$ under ORACLE on Allen--Cahn, and $11$ versus $0$ on Wave (Fig.~\ref{fig:geometric_analysis}c). The Euclidean ablation therefore produces substantially more candidates whose apparent local improvement does not persist on held-out constraints. This diagnostic links the geometry used
to construct the alignment with its off-support reliability.
Figure~\ref{fig:geometric_analysis}(d) shows the resulting endpoint error. Removing the authority bound gives the lowest Allen-Cahn error, but provides a smaller gain on Wave; full ORACLE improves both PDEs substantially. We therefore do not interpret every mechanism as independently necessary. Rather, ORACLE combines three roles: the native optimizer geometry determines
\emph{how} alignment is measured, the authority bound limits \emph{how far} the proposal may be redirected, and held-out validation determines \emph{whether} the aligned proposal is committed. Extended mechanism analysis appears in Appendix~\ref{app:geometric_analysis}.

\vspace{-0.2cm}
\section{Conclusion and Limitations}
\vspace{-0.2cm}

We introduced ORACLE, which changes both where and how constraints interact
with optimization: it acts on the realized native update, constructs a joint
local model of heterogeneous constraints at the proposed endpoint, and aligns
the resulting update in the geometry of the optimizer that produced the step. Across five experimental questions, ORACLE improves or matches Native in 94\% of optimizer-PDE settings and 92\% of model-formulation settings, with endpoint
error reductions of up to 90.6\% and improvements that persist across the solution domain and physical time.  These improvements come with a clear trade-off: ORACLE incurs additional computation and depends on the fidelity of its local correction model. Future work will target cheaper, selective interventions through adaptive correction budgets and reuse of native optimizer geometry, together with richer local models for strongly nonlinear regimes. Extending the current single-step analysis to trajectory-level guarantees and testing optimizer-relative alignment beyond PDE learning are natural next steps.

\section*{Acknowledgements}
This work was supported by CRADA22-0034-J003 and and NVIDIA Academic Grant Program.

\bibliography{iclr2027_conference}
\bibliographystyle{iclr2027_conference}

\appendix
\appendix

\section{Extended Related Work}
\label{app:related_work}

The main text positions ORACLE relative to the principal families of
constraint-handling methods. Here we expand that discussion around the
interface central to this work: the distinction between the gradient supplied
to an optimizer and the parameter displacement that the optimizer ultimately
produces.

\subsection{Optimizer Geometry and the Realized Update}

Modern optimizers transform the instantaneous gradient using information
accumulated across training. AdaGrad adapts coordinate scales from gradient
history~\cite{ward2020adagrad}; Adam combines first- and second-moment
statistics~\cite{kingma2014adam}; and Adafactor factorizes adaptive statistics
to reduce memory~\cite{shazeer2018adafactor}. More structured methods alter
the update through non-diagonal geometry: natural gradient uses the Fisher
metric~\cite{kakade2001natural}, K-FAC approximates Fisher blocks with
Kronecker factors~\cite{grosse2016kronecker}, Shampoo maintains
tensor-structured preconditioners~\cite{gupta2018shampoo}, and SOAP combines
structured preconditioning with Adam-style adaptation in an evolving basis
~\cite{vyas2025soap}. Newton, Gauss Newton, and quasi-Newton methods likewise
reshape parameter motion using curvature information
~\cite{berahas2016multi}.

Across these methods, the gradient is an input to the optimizer rather than
the parameter movement itself. Momentum, adaptive scaling, and structured
preconditioning can change both the scale and direction of the resulting
step. ORACLE refers to this optimizer-transformed displacement as the
\emph{realized optimizer update}. This distinction becomes consequential when
constraint information must interact with an optimizer whose state already
defines a nontrivial geometry.

\subsection{Constraint Handling and the Intervention Point}

Classical constrained optimization typically introduces constraint
information before or during construction of the optimization trajectory.
Penalty methods add violation terms to the objective; primal dual methods
jointly evolve parameter and multiplier variables~\cite{nandwani2019primal};
and augmented-Lagrangian approaches combine multiplier dynamics with explicit
feasibility penalties~\cite{dener2020training,lavado2023achieving}.
Related formulations extend these mechanisms to constrained neural
networks~\cite{fioretto2020lagrangian}. In each case, constraint information
affects the signal from which the optimizer subsequently constructs its step.

Multi-objective methods intervene more directly on gradients. GradNorm
balances objectives using gradient magnitudes~\cite{chen2018gradnorm}; MGDA
constructs a common descent direction~\cite{desideri2012multiple}; PCGrad
removes pairwise conflicting components~\cite{yu2020gradient}; CAGrad trades
average descent against worst-objective conflict~\cite{liu2021conflict}; and
Nash-MTL derives a shared direction through bargaining
~\cite{navon2022multi}. These methods address imbalance or directional
conflict before the native optimizer applies its state-dependent
transformation.

Projection and local constrained-optimization methods act closer to the
parameter movement itself. Projected-gradient methods map candidate steps
toward feasible sets, SQP constructs locally constrained subproblems, and
trust-region mechanisms restrict movement to regions where local models remain
reliable. Related learning methods include GEM, which projects gradients that
would interfere with previously learned tasks~\cite{lopez2017gradient}, and
differentiable projection layers that explicitly map predictions or parameters
toward feasibility~\cite{huang2021differentiable}. These methods separate a
proposed direction from a constraint-induced modification, but the proposal is
often defined before a stateful optimizer has produced its realized update.

This ordering matters when the optimizer itself carries useful structure.
For Adam, SOAP, Shampoo, or other stateful optimizers, modifying the raw
gradient does not directly specify the parameter displacement that will follow.
ORACLE instead takes the realized native step as the reference movement and
asks whether that movement should be aligned with the constraints.

\subsection{Constraint Geometry in Physics-Informed Learning}

PINNs provide a demanding setting for this distinction because their training
objectives combine PDE residuals, initial conditions, boundary conditions, and
possibly observations~\cite{raissi2019physics}. These components can differ
substantially in scale, conditioning, spatial support, and gradient direction.

Several methods address this difficulty before the optimizer forms its step.
Gradient-statistics-based annealing adapts loss weights from constituent
gradient magnitudes~\cite{wang2021understanding}; NTK-based weighting accounts
for unequal convergence rates~\cite{wang2021eigenvector}; ReLoBRaLo adapts
weights using relative loss progress~\cite{bischof2025multi}; and MultiAdam
maintains separate adaptive statistics for different PINN loss components
~\cite{yao2023multiadam}. Other work instead strengthens the optimizer
geometry itself through curvature-aware or structured methods, including
NysNewton-CG~\cite{rathore2024challenges}, Shampoo
~\cite{gupta2018shampoo}, and SOAP~\cite{vyas2025soap}. Recent analyses further
show that PINN difficulty is not explained by gradient magnitude alone:
constituent gradients can remain directionally incompatible even after loss
balancing~\cite{wang2026gradient}, while benchmark studies show substantial
variation in optimizer behavior across PDE families~\cite{hao2024pinnacle}.

These developments create the setting studied by ORACLE. The native optimizer
may already encode useful task-dependent scaling or preconditioning, while the
physical constraints provide a separate collection of local sensitivities.
Combining the two before optimization changes the signal from which the native
update is constructed; operating only on the raw gradient acts before the
optimizer geometry has been expressed. ORACLE instead preserves the realized
native update as the reference movement.

\subsection{Positioning ORACLE}

ORACLE operates after the native optimizer has formed its realized step. It first recovers and certifies the geometry associated with that step, then evaluates whether the movement is locally adverse to a sampled constraint model. When alignment is required, ORACLE constructs the adjustment in the
same optimizer-relative metric, bounds its authority relative to the native step, and commits it only after nonlinear validation and evaluation on independent support.

The distinction is therefore not merely \emph{when} constraint information is introduced. ORACLE separates two forms of structure: \emph{optimizer geometry}, which determines how the native optimizer prefers to move, and \emph{constraint geometry}, which determines how that realized movement affects local constraint violation. Matrix-free Jacobian vector and
vector Jacobian products allow the latter to be represented without materializing the parameter-space Jacobian, while native-relative authority control and held-out validation regulate whether the resulting alignment is committed.

ORACLE therefore complements rather than replaces the underlying optimizer or constraint model. Its contribution is the interface between them: constraints are evaluated against the movement the optimizer actually produced, and any alignment is expressed relative to the geometry of that same movement.

\subsection{Connecting ORACLE to different fields}

ORACLE shifts constraint intervention from the raw gradient to the realized parameter update, computing a matrix-free, bounded alignment strictly within the native optimizer's certified geometry. This post-optimizer architecture seamlessly integrates into parametric constrained optimization frameworks like DePICT by ensuring decision-preserving interfaces are not distorted by the underlying preconditioner. By evaluating constraints at the proposed endpoint rather than the objective~\cite{grover2026depictdecisionpreservinginterfaceconstrained}, it enables unlearning models ~\cite{ranjan2026razor, bourtoule2021machine, ranjan2026g,cao2015towards} to maximize target forgetting while independently treating retention as an endpoint geometric constraint. Additionally, it provides a critical safeguard for multi-agent systems and Vision-Language-Action (VLA) models by explicitly blocking adverse policy steps into unsafe states~\cite{kim2024openvla, brohan2023rt, grover2026embodied}. Furthermore, because ORACLE dynamically recovers the metric of stateful optimizers before intervening, which is currently being explored in modern safe reinforcement learning ~\cite{gu2024review, yang2026cbf, tayal2026safe, ranjan2026persa}, it naturally extends to physics-informed deep learning tasks relying on advanced second-order algorithms like SOAP, Muon ensuring boundary and initial conditions are satisfied without disrupting the optimizer's core convergence properties~\cite{vyas2025soap, du2026newton}.
\section{Supporting Theory and Derivations}
\label{app:theory}

This appendix provides the technical details supporting
Section~\ref{sec:method}. We first formalize the finite-support constraint
estimator, then derive the optimizer-specific geometry adapters used to recover
$P_t$, characterize when the PSGD--Kron adapter must abstain, and justify the
endpoint signal used to open the ORACLE alignment path. Subsequent subsections
derive the optimizer-relative alignment, authority bound, and validation
mechanisms.

\subsection{Finite-Support Constraint Estimator}
\label{app:sampling_unbiasedness}

Fix iteration $t$ and let $\mathcal{F}_t$ contain all randomness determined
before the construction support $S_t$ is drawn. In particular,
$\theta_t$, the realized native step $s_t$, the proposed endpoint
$\bar{\theta}_t=\theta_t+s_t$, the family weights $\omega_{k,t}$, and the
sample counts $m_k$ are $\mathcal{F}_t$-measurable. For each constraint family
$k$, draw
\[
z_{k,1},\ldots,z_{k,m_k}\overset{\mathrm{i.i.d.}}{\sim}\nu_k.
\]

The sampled constraint energy is
\begin{equation}
\Phi_{S_t}(\theta)
=
\sum_{k=1}^{K}
\omega_{k,t}
\frac{1}{m_k}
\sum_{j=1}^{m_k}
r_k(\theta;z_{k,j})^2.
\label{eq:app_sampled_energy}
\end{equation}
This is equivalent to the weighted residual form
$\frac12 r_{S_t}(\theta)^\top W_t r_{S_t}(\theta)$ used in the main text,
with each residual from family $k$ assigned weight
$2\omega_{k,t}/m_k$.

For any $\mathcal{F}_t$-measurable evaluation point $\Theta_t$,
\begin{align}
\mathbb{E}
\left[
\Phi_{S_t}(\Theta_t)
\mid
\mathcal{F}_t
\right]
&=
\sum_{k=1}^{K}
\omega_{k,t}
\mathbb{E}_{z\sim\nu_k}
\left[
r_k(\Theta_t;z)^2
\right] \\
&=
\Phi_t(\Theta_t).
\end{align}
Hence
\begin{equation}
\boxed{
\mathbb{E}
\left[
\Phi_{S_t}(\Theta_t)
\mid
\mathcal{F}_t
\right]
=
\Phi_t(\Theta_t)
}
\label{eq:app_unbiased_energy}
\end{equation}
and, whenever differentiation and expectation may be interchanged,
\begin{equation}
\mathbb{E}
\left[
\nabla\Phi_{S_t}(\Theta_t)
\mid
\mathcal{F}_t
\right]
=
\nabla\Phi_t(\Theta_t).
\label{eq:app_unbiased_gradient}
\end{equation}

The ordering used by ORACLE satisfies this condition because
$\bar{\theta}_t$ is fixed before $S_t$ is sampled. For finite stored
constraint sets, $\nu_k$ denotes the corresponding empirical distribution.

\subsection{Certified Optimizer Geometry}
\label{app:optimizer_adapters}

Let $u_t$ denote the optimizer's effective direction after its stateful
transformation but before application of its local mobility. ORACLE uses an
optimizer-relative geometry only when it can identify an SPD operator
$B_t$ satisfying
\begin{equation}
s_t=-B_tu_t,
\qquad
P_t=B_t^{-1}\succ0,
\label{eq:app_metric_relation}
\end{equation}
and this representation reconstructs the realized deployed step.

\paragraph{Certification.}
Define the reconstruction defect
\[
e_t=s_t+B_tu_t.
\]
ORACLE accepts the recovered geometry only if
\begin{equation}
\boxed{
\lVert e_t\rVert_2
\le
\tau_{\mathrm{rel}}\lVert s_t\rVert_2
+
\tau_{\mathrm{abs}},
\qquad
\tau_{\mathrm{rel}}=10^{-4},
\quad
\tau_{\mathrm{abs}}=10^{-9}.
}
\label{eq:app_cert_gate}
\end{equation}
If this test or an optimizer-specific validity condition fails, ORACLE
abstains and preserves the Native proposal; it never substitutes
$P_t=I$.

For nonzero native steps, the corresponding relative reconstruction residual is
\begin{equation}
\varepsilon_{\mathrm{cert},t}
=
\frac{
\lVert s_t+B_tu_t\rVert_2
}{
\lVert s_t\rVert_2
}.
\label{eq:app_cert_residual}
\end{equation}
Runtime certification is evaluated on the deployed numerical path; exact
algebraic identities alone are not treated as production certification.

\paragraph{SGD.}
For SGD without momentum,
\[
s_t=-\eta_t g_t.
\]
Thus
\begin{equation}
u_t=g_t,
\qquad
B_t=\eta_t I,
\qquad
P_t=\eta_t^{-1}I.
\end{equation}
For $\eta_t>0$, both operators are SPD and
\eqref{eq:app_metric_relation} holds exactly.

\paragraph{Adam.}
Let $\widehat\mu_t$ and $\widehat\nu_t$ denote the bias-corrected moments used
by the deployed Adam update, and define
\[
D_t
=
\operatorname{diag}
\left(
\sqrt{\widehat\nu_t+\varepsilon_{\mathrm{root}}}
+\varepsilon
\right)
\succ0.
\]
Then
\[
s_t=-\eta_tD_t^{-1}\widehat\mu_t,
\]
so
\begin{equation}
u_t=\widehat\mu_t,
\qquad
B_t=\eta_tD_t^{-1},
\qquad
P_t=\eta_t^{-1}D_t.
\end{equation}
The adapter follows the numerical stabilizers and state timing of the deployed
optimizer exactly.

\paragraph{SOAP.}
Let $Q_t$ denote the orthogonal basis that generated the realized SOAP step,
$D_t\succ0$ its adaptive denominator, $h_t>0$ the scalar step factor, and
$u_t^{\mathrm{soap}}$ the effective momentum in parameter coordinates. The
realized step is
\[
s_t
=
-h_tQ_t^\top D_t^{-1}Q_tu_t^{\mathrm{soap}},
\]
which gives
\begin{equation}
u_t=u_t^{\mathrm{soap}},
\qquad
B_t=h_tQ_t^\top D_t^{-1}Q_t,
\qquad
P_t=h_t^{-1}Q_t^\top D_tQ_t.
\end{equation}
Because $Q_t$ is orthogonal and $D_t\succ0$, both $B_t$ and $P_t$ are SPD.

The basis timing is essential. If SOAP refreshes its basis after constructing
the current step, certification must use the basis that actually generated
$s_t$; using the refreshed basis describes a different operator and may fail
the reconstruction test.

\paragraph{PSGD--Kron.}
In the unclipped linear regime, let
\[
K_t^{(\ell)}
=
\bigotimes_i A_{i,t}^{(\ell)},
\qquad
A_{i,t}^{(\ell)}\succ0,
\]
and
\[
K_t
=
\bigoplus_\ell K_t^{(\ell)}.
\]
Kronecker products and direct sums of SPD operators remain SPD. If
\[
s_t=-\eta_tK_tu_t^{\mathrm{kron}},
\]
then
\begin{equation}
u_t=u_t^{\mathrm{kron}},
\qquad
B_t=\eta_tK_t,
\qquad
P_t=\eta_t^{-1}K_t^{-1}.
\end{equation}
This representation is valid only while the deployed update remains linear in
the preconditioned direction.

\subsection{PSGD--Kron Clipping and Abstention}
\label{app:kron_clipping}

Let
\[
p=K_tu_t^{\mathrm{kron}}\in\mathbb{R}^{n}
\]
denote the linear preconditioned direction before clipping. The deployed
implementation first applies the global normalization
\begin{equation}
\gamma(p)
=
p\,
\frac{\sqrt n}
{\max(\sqrt n,\lVert p\rVert_2)}
\end{equation}
and then the elementwise map
\begin{equation}
h(y)_j
=
\min\{1,\max\{-1,y_j\}\}.
\end{equation}
The realized step is therefore
\[
s_t=-\eta_t h(\gamma(p)).
\]

A global norm rescaling alone would preserve an SPD linear mobility. In the
deployed rule, however,
\[
\lVert p\rVert_2>\sqrt n
\quad\Longrightarrow\quad
\lVert p\rVert_\infty
\ge
\frac{\lVert p\rVert_2}{\sqrt n}
>1,
\]
and hence
\begin{equation}
\boxed{
\text{global clipping active}
\quad\Longrightarrow\quad
\text{elementwise clipping active}.
}
\label{eq:kron_clip_implication}
\end{equation}

Once the elementwise clip binds, the optimizer response is coordinatewise
nonlinear. The linear Kronecker operator can no longer represent the mobility
used to evaluate arbitrary alignment directions. ORACLE therefore rejects the
Kron adapter on such steps rather than replacing it with an unverified
geometry.

\subsection{Endpoint Constraint Signal}
\label{app:endpoint_signal}

Suppress the iteration index and write
\[
\theta=\theta_t,
\qquad
s=s_t,
\qquad
\bar{\theta}=\theta+s,
\qquad
\Phi=\Phi_{S_t},
\qquad
c=\nabla\Phi(\bar{\theta}),
\qquad
P=P_t.
\]

Assume that $\nabla\Phi$ is Lipschitz along the realized native segment
$\{\theta+\tau s:\tau\in[0,1]\}$ in the optimizer metric:
\begin{equation}
\lVert
\nabla\Phi(x)-\nabla\Phi(y)
\rVert_{P^{-1}}
\le
L_\Phi^P
\lVert x-y\rVert_P.
\label{eq:app_metric_smoothness}
\end{equation}
By the fundamental theorem of calculus,
\begin{align}
\Phi(\bar{\theta})-\Phi(\theta)-c^\top s
&=
\int_0^1
\left[
\nabla\Phi(\theta+\tau s)
-
\nabla\Phi(\bar{\theta})
\right]^\top
s\,d\tau .
\end{align}
Applying dual Cauchy--Schwarz and
\eqref{eq:app_metric_smoothness} yields
\begin{equation}
\boxed{
\left|
\Phi_{S_t}(\bar{\theta}_t)
-
\Phi_{S_t}(\theta_t)
-
c_t^\top s_t
\right|
\le
\frac{L_{\Phi,t}^{P}}{2}
\lVert s_t\rVert_{P_t}^2 .
}
\label{eq:app_endpoint_bound}
\end{equation}
Thus $c_t^\top s_t$ is the first-order term associated with the sampled
constraint-energy change along the realized native movement. Because ORACLE
does not estimate $L_{\Phi,t}^{P}$ at runtime, the deployed rule
\[
c_t^\top s_t>0
\]
is a first-order trigger, not an unconditional finite-step sign certificate.

\paragraph{Why evaluate at the endpoint?}
Define
\[
h(\tau)=\Phi(\theta+\tau s).
\]
If $\Phi$ is convex along the realized native segment, then $h'$ is
nondecreasing and
\[
h'(0)
\le
h(1)-h(0)
\le
h'(1).
\]
Equivalently,
\begin{equation}
\boxed{
\nabla\Phi(\theta)^\top s
\le
\Phi(\theta+s)-\Phi(\theta)
\le
\nabla\Phi(\theta+s)^\top s.
}
\label{eq:app_endpoint_inequality}
\end{equation}
Therefore,
\begin{equation}
\nabla\Phi(\theta+s)^\top s\le0
\quad\Longrightarrow\quad
\Phi(\theta+s)\le\Phi(\theta)
\end{equation}
under segment convexity.

The endpoint signal is therefore asymmetric. A non-positive value provides a
one-sided condition for preserving the Native proposal under the stated
assumption, whereas a positive value only opens the alignment path. The
resulting candidate must still satisfy the alignment-descent, authority,
nonlinear-agreement, and held-out validation conditions before commitment.

\subsection{Optimizer-Relative Alignment}
\label{app:alignment_derivation}

This subsection derives the optimizer-relative alignment used in
Section~\ref{sec:alignment_commit}. We establish the equivalence between its
parameter- and residual-space forms, show that both exact and finite-CG
solutions produce first-order constraint-descent directions on triggered
steps, and give the matrix-free operator used in implementation.

\paragraph{Parameter- and residual-space formulations.}
At the proposed endpoint $\bar{\theta}_t$, define
\[
    C_t=W_t^{1/2}J_t,
    \qquad
    b_t=W_t^{1/2}r_t.
\]
Before authority control, ORACLE computes
\begin{equation}
    \widehat d_t
    =
    \arg\min_d
    \left\{
        \frac12\lVert b_t+C_td\rVert_2^2
        +
        \frac{\lambda_t}{2}\lVert d\rVert_{P_t}^2
    \right\},
    \qquad
    \lambda_t>0.
    \label{eq:app_alignment_primal}
\end{equation}
Because $P_t\succ0$, the objective is strictly convex. Its optimality
condition is
\begin{equation}
    \left(C_t^\top C_t+\lambda_tP_t\right)\widehat d_t
    =
    -C_t^\top b_t
    =
    -c_t.
    \label{eq:app_alignment_normal}
\end{equation}

Define the optimizer-relative residual Gram operator
\begin{equation}
    G_t
    =
    C_tP_t^{-1}C_t^\top
    =
    W_t^{1/2}J_tP_t^{-1}J_t^\top W_t^{1/2}
    \succeq0.
    \label{eq:app_alignment_gram}
\end{equation}
If
\begin{equation}
    (G_t+\lambda_tI)z_t=b_t,
    \label{eq:app_alignment_residual_system}
\end{equation}
then
\begin{equation}
    \boxed{
    \widehat d_t
    =
    -P_t^{-1}C_t^\top z_t
    =
    -P_t^{-1}J_t^\top W_t^{1/2}z_t.
    }
    \label{eq:app_alignment_candidate}
\end{equation}
To verify the equivalence,
\begin{align}
    &
    \left(C_t^\top C_t+\lambda_tP_t\right)
    \left(-P_t^{-1}C_t^\top z_t\right)
    \nonumber\\
    &\qquad
    =
    -C_t^\top
    \left(
        C_tP_t^{-1}C_t^\top+\lambda_tI
    \right)z_t
    \\
    &\qquad
    =
    -C_t^\top(G_t+\lambda_tI)z_t
    =
    -C_t^\top b_t
    =
    -c_t.
\end{align}
Hence \eqref{eq:app_alignment_candidate} satisfies
\eqref{eq:app_alignment_normal}.

\paragraph{Exact-solve descent.}
Since $c_t=C_t^\top b_t$,
\begin{equation}
    -c_t^\top\widehat d_t
    =
    b_t^\top
    G_t(G_t+\lambda_tI)^{-1}b_t.
    \label{eq:app_exact_alignment_descent}
\end{equation}
Let
\[
    G_t
    =
    Q\operatorname{diag}(\sigma_i)Q^\top,
    \qquad
    \beta_i=q_i^\top b_t,
    \qquad
    \sigma_i\ge0.
\]
Then
\begin{equation}
    -c_t^\top\widehat d_t
    =
    \sum_i
    \frac{\sigma_i}{\sigma_i+\lambda_t}\beta_i^2.
    \label{eq:app_exact_alignment_spectral}
\end{equation}
Moreover,
\begin{equation}
    b_t^\top G_tb_t
    =
    c_t^\top P_t^{-1}c_t
    =
    \lVert c_t\rVert_{P_t^{-1}}^2.
    \label{eq:app_alignment_gram_identity}
\end{equation}

On a triggered step, $c_t^\top s_t>0$, so $c_t\neq0$. Therefore
$b_t^\top G_tb_t>0$, implying that $b_t$ has nonzero mass on at least one
eigenspace with $\sigma_i>0$. Equation~\ref{eq:app_exact_alignment_spectral}
is therefore strictly positive, and
\begin{equation}
    \boxed{
        c_t^\top\widehat d_t<0.
    }
    \label{eq:app_exact_alignment_strict}
\end{equation}
Thus the exact optimizer-relative solution is a strict first-order
constraint-descent direction whenever the alignment path is triggered.

\paragraph{Finite-CG descent.}
The deployed method solves the residual system with conjugate gradients
rather than forming $(G_t+\lambda_tI)^{-1}$. Let
\[
    A_t=G_t+\lambda_tI\succ0,
\]
and let $z_{t,k}$ be the $k$-th CG iterate for
$A_tz=b_t$, initialized at $z_{t,0}=0$. Define
\begin{equation}
    \widehat d_{t,k}
    =
    -P_t^{-1}C_t^\top z_{t,k}.
    \label{eq:app_cg_alignment_candidate}
\end{equation}

Let
\[
    \delta_{t,k}
    =
    b_t-A_tz_{t,k}
\]
denote the CG residual. For $k\ge1$, CG orthogonality gives
\[
    \delta_{t,k}
    \perp
    \mathcal{K}_k(A_t,b_t),
\]
while both $b_t$ and $z_{t,k}$ belong to
$\mathcal{K}_k(A_t,b_t)$. Hence
\begin{align}
    b_t^\top A_tz_{t,k}
    &=
    b_t^\top(b_t-\delta_{t,k})
    =
    \lVert b_t\rVert_2^2,
    \label{eq:app_cg_b_orthogonality}
    \\
    b_t^\top z_{t,k}
    &=
    (A_tz_{t,k}+\delta_{t,k})^\top z_{t,k}
    =
    \lVert z_{t,k}\rVert_{A_t}^2.
    \label{eq:app_cg_z_orthogonality}
\end{align}

Using $G_t=A_t-\lambda_tI$,
\begin{align}
    -c_t^\top\widehat d_{t,k}
    &=
    b_t^\top G_tz_{t,k}
    \\
    &=
    \lVert b_t\rVert_2^2
    -
    \lambda_t\lVert z_{t,k}\rVert_{A_t}^2.
    \label{eq:app_cg_alignment_identity}
\end{align}

From \eqref{eq:app_cg_z_orthogonality},
\[
    \lVert z_{t,k}\rVert_{A_t}^2
    =
    \left\langle
        A_t^{-1}b_t,\,
        z_{t,k}
    \right\rangle_{A_t}.
\]
Cauchy--Schwarz therefore gives
\begin{equation}
    \lVert z_{t,k}\rVert_{A_t}^2
    \le
    b_t^\top A_t^{-1}b_t.
    \label{eq:app_cg_alignment_norm_bound}
\end{equation}

In the eigenbasis of $G_t$,
\begin{equation}
    \lambda_t b_t^\top A_t^{-1}b_t
    =
    \sum_i
    \frac{\lambda_t}{\sigma_i+\lambda_t}\beta_i^2.
    \label{eq:app_alignment_resolvent}
\end{equation}
The coefficient equals one only when $\sigma_i=0$ and is strictly smaller
than one when $\sigma_i>0$. By
\eqref{eq:app_alignment_gram_identity}, a triggered step has nonzero
$\beta_i$ for at least one $\sigma_i>0$. Hence
\begin{equation}
    \lambda_t b_t^\top A_t^{-1}b_t
    <
    \lVert b_t\rVert_2^2.
    \label{eq:app_alignment_resolvent_strict}
\end{equation}
Combining
\eqref{eq:app_cg_alignment_identity},
\eqref{eq:app_cg_alignment_norm_bound}, and
\eqref{eq:app_alignment_resolvent_strict} yields
\begin{equation}
    \boxed{
        c_t^\top\widehat d_{t,k}<0,
        \qquad
        k\ge1.
    }
    \label{eq:app_cg_alignment_strict}
\end{equation}

Thus every nonzero finite-CG iterate is a strict first-order
constraint-descent direction on a triggered step in exact arithmetic. The
runtime sign test in Algorithm~\ref{alg:oracle} therefore serves as a
finite-precision safeguard rather than an additional modeling assumption.

\paragraph{Relative damping.}
ORACLE scales the absolute damping to the residual geometry:
\begin{equation}
    \lambda_t
    =
    \lambda_{\mathrm{rel},t}\varrho_t,
    \qquad
    \varrho_t
    =
    \frac{b_t^\top G_tb_t}{b_t^\top b_t}.
    \label{eq:app_relative_damping_rq}
\end{equation}
On a triggered step, $c_t\neq0$, and therefore $b_t\neq0$, so the quotient
is well defined. Using \eqref{eq:app_alignment_gram_identity} and
\[
    b_t^\top b_t
    =
    2\Phi_{S_t}(\bar{\theta}_t),
\]
we obtain
\begin{equation}
    \boxed{
    \varrho_t
    =
    \frac{
        \lVert c_t\rVert_{P_t^{-1}}^2
    }{
        2\Phi_{S_t}(\bar{\theta}_t)
    }.
    }
    \label{eq:app_relative_damping_identity}
\end{equation}
Equivalently, $\varrho_t$ is the Rayleigh quotient of $G_t$ along the
current weighted-residual direction. It therefore provides a
residual-dependent scale for the regularization rather than using a fixed
absolute damping across problems and optimizer geometries. The deployed
controller for $\lambda_{\mathrm{rel},t}$ is specified in
Section~\ref{sec:method}.

\paragraph{Matrix-free application.}
Neither $J_t$ nor $G_t$ is materialized. For any residual-space vector $v$,
\begin{equation}
    \boxed{
    G_tv
    =
    W_t^{1/2}J_t
    \left[
        P_t^{-1}
        \left(
            J_t^\top W_t^{1/2}v
        \right)
    \right].
    }
    \label{eq:app_matrix_free_gram}
\end{equation}
Each application of $G_t$ therefore requires one VJP through
$J_t^\top$, one application of the certified mobility $P_t^{-1}$, and one
JVP through $J_t$. CG can consequently solve
\eqref{eq:app_alignment_residual_system} without forming either the
residual Jacobian or a parameter-space matrix.
\subsection{Authority, Validation, and Exact Native Reduction}
\label{app:authority_validation}

This subsection supports the authority and validation mechanisms in
Section~\ref{sec:method}. We establish the native-relative authority bound,
derive its geometric interpretation, formalize the nonlinear-agreement and
held-out validation tests, and show that any rejected alignment recovers the
exact Native update from the current optimizer state.

\paragraph{Native-relative alignment authority.}
On a triggered step, define
\[
    h_t=c_t^\top s_t>0,
    \qquad
    \kappa_t=-c_t^\top\widehat d_t>0.
\]
Here $h_t$ measures the detected first-order constraint increase along the
Native step, while $\kappa_t$ measures the first-order decrease provided by
the optimizer-relative alignment direction.

ORACLE defines
\begin{equation}
    \alpha_{\mathrm{need},t}
    =
    \frac{h_t}
         {\kappa_t+\varepsilon_{\mathrm{auth}}},
    \qquad
    \alpha_{\mathrm{safe},t}
    =
    \frac{
        \beta_t\lVert s_t\rVert_{P_t}
    }{
        \lVert\widehat d_t\rVert_{P_t}
    },
    \label{eq:app_authority_scales}
\end{equation}
with $\varepsilon_{\mathrm{auth}}=10^{-30}$, and sets
\begin{equation}
    \alpha_t
    =
    \min\{
        \alpha_{\mathrm{need},t},
        \alpha_{\mathrm{safe},t}
    \},
    \qquad
    \widetilde d_t
    =
    \alpha_t\widehat d_t.
    \label{eq:app_authority_scale}
\end{equation}

Without the numerical guard,
$\alpha_{\mathrm{need},t}=h_t/\kappa_t$ is exactly the scale required to
neutralize the detected first-order constraint signal:
\[
    c_t^\top
    \left(
        s_t+\alpha_{\mathrm{need},t}\widehat d_t
    \right)
    =
    h_t-\alpha_{\mathrm{need},t}\kappa_t
    =
    0.
\]
The numerical guard keeps the deployed scale infinitesimally below this exact
cancellation point.

Since
$\alpha_t\le\alpha_{\mathrm{safe},t}$,
\begin{align}
    \lVert\widetilde d_t\rVert_{P_t}
    &=
    \alpha_t
    \lVert\widehat d_t\rVert_{P_t}
    \\
    &\le
    \beta_t
    \lVert s_t\rVert_{P_t}.
\end{align}
Therefore,
\begin{equation}
    \boxed{
    \lVert\widetilde d_t\rVert_{P_t}
    \le
    \beta_t
    \lVert s_t\rVert_{P_t}.
    }
    \label{eq:app_bounded_authority}
\end{equation}
The denominator of $\alpha_{\mathrm{safe},t}$ is nonzero on the triggered
alignment path because
$c_t^\top\widehat d_t<0$ implies $\widehat d_t\neq0$.

\paragraph{Geometric interpretation of the authority bound.}
Let
\[
    x=P_t^{1/2}s_t,
    \qquad
    y=P_t^{1/2}\widetilde d_t.
\]
Equation~\ref{eq:app_bounded_authority} implies
\[
    \lVert y\rVert_2
    \le
    \beta_t\lVert x\rVert_2.
\]
Moreover, the $P_t$-angle between $s_t$ and
$s_t+\widetilde d_t$ equals the Euclidean angle between $x$ and $x+y$.

Normalize $\lVert x\rVert_2=1$ and decompose
\[
    y=ax+bq,
    \qquad
    q^\top x=0,
    \qquad
    \lVert q\rVert_2=1,
    \qquad
    a^2+b^2\le\beta_t^2.
\]
If
\[
    \vartheta
    =
    \angle(x,x+y),
\]
then
\begin{equation}
    \sin^2\vartheta
    =
    \frac{b^2}
         {(1+a)^2+b^2}.
    \label{eq:app_angle_sine}
\end{equation}
For fixed $a$, this quantity increases with $b^2$, so its maximum lies on
$a^2+b^2=\beta_t^2$. Substituting
$b^2=\beta_t^2-a^2$ gives
\[
    f(a)
    =
    \frac{\beta_t^2-a^2}
         {1+2a+\beta_t^2}.
\]
The interior maximizer is $a=-\beta_t^2$, for which
\[
    f(-\beta_t^2)
    =
    \beta_t^2.
\]
Hence, for $0\le\beta_t<1$,
\begin{equation}
    \boxed{
    \angle_{P_t}
    \left(
        s_t,\,
        s_t+\widetilde d_t
    \right)
    \le
    \arcsin(\beta_t).
    }
    \label{eq:app_authority_angle}
\end{equation}

The bound is sharp. Equality is attained in transformed coordinates by
\[
    y
    =
    -\beta_t^2x
    +
    \beta_t\sqrt{1-\beta_t^2}\,q,
    \qquad
    q\perp x.
\]
Thus $\beta_t$ has a direct geometric interpretation: it limits how far the
aligned movement may rotate away from the direction selected by the Native
optimizer, measured in that optimizer's own metric.

\paragraph{Nonlinear agreement.}
The alignment direction is constructed from a local linearization, so ORACLE
checks whether the predicted reduction agrees with the true nonlinear
constraint energy before proceeding.

On the same construction support $S_t$, define
\begin{align}
    \mathrm{Pred}_t
    &=
    \Phi_{S_t}(\bar{\theta}_t)
    -
    \frac12
    \left\lVert
        b_t+C_t\widetilde d_t
    \right\rVert_2^2,
    \label{eq:app_predicted_reduction}
    \\
    \mathrm{Ared}_t
    &=
    \Phi_{S_t}(\bar{\theta}_t)
    -
    \Phi_{S_t}
    \left(
        \bar{\theta}_t+\widetilde d_t
    \right).
    \label{eq:app_actual_reduction}
\end{align}
Recall that
$C_t=W_t^{1/2}J_t$ and $b_t=W_t^{1/2}r_t$.

The candidate proceeds only if
\begin{equation}
    \mathrm{Pred}_t>0,
    \qquad
    \mathrm{Ared}_t>0,
    \qquad
    \rho_t
    =
    \frac{\mathrm{Ared}_t}
         {\mathrm{Pred}_t}
    \ge
    \rho_{\min}.
    \label{eq:app_pred_ared_gate}
\end{equation}
Because $S_t$ is used both to construct and evaluate the candidate, this test
measures agreement between the local model and the nonlinear constraints; it
does not provide independent evidence that the alignment improves unseen
constraint samples.

\paragraph{Independent held-out validation.}
ORACLE therefore fixes $\widetilde d_t$ before drawing an independent audit
support $S'_t$. Let $\mathcal{G}_t$ contain all information available after
the candidate is fixed but before $S'_t$ is sampled. Conditional on
$\mathcal{G}_t$, the audit samples are independent draws from their prescribed
constraint-family distributions.

For family $k$, define
\begin{equation}
    X_{k,t}(z)
    =
    \omega_{k,t}
    \left[
        r_k(\bar{\theta}_t+\widetilde d_t;z)^2
        -
        r_k(\bar{\theta}_t;z)^2
    \right].
    \label{eq:app_audit_sample_difference}
\end{equation}
With $m'_k$ audit samples from family $k$, the empirical change is
\begin{equation}
    \widehat{\Delta}^{\,\mathrm{audit}}_t
    =
    \sum_{k=1}^{K}
    \frac{1}{m'_k}
    \sum_{j=1}^{m'_k}
    X_{k,t}(z'_{k,j}).
    \label{eq:app_audit_estimator}
\end{equation}
The corresponding population change is
\begin{align}
    \Delta_t^{\mathrm{pop}}
    &=
    \sum_{k=1}^{K}
    \mathbb{E}
    \left[
        X_{k,t}(z)
        \mid
        \mathcal{G}_t
    \right]
    \nonumber\\
    &=
    \Phi_t
    \left(
        \bar{\theta}_t+\widetilde d_t
    \right)
    -
    \Phi_t(\bar{\theta}_t).
    \label{eq:app_population_change}
\end{align}
Therefore,
\begin{equation}
    \boxed{
    \mathbb{E}
    \left[
        \widehat{\Delta}^{\,\mathrm{audit}}_t
        \mid
        \mathcal{G}_t
    \right]
    =
    \Delta_t^{\mathrm{pop}}.
    }
    \label{eq:app_audit_unbiased}
\end{equation}

Assume that, conditional on $\mathcal{G}_t$, each
$X_{k,t}(z)$ lies in an interval of width $R_{k,t}$. Hoeffding's inequality
then gives, for any $u>0$,
\begin{equation}
    \Pr\!\left[
        \widehat{\Delta}^{\,\mathrm{audit}}_t
        -
        \Delta_t^{\mathrm{pop}}
        \le -u
        \,\middle|\,
        \mathcal{G}_t
    \right]
    \le
    \exp\!\left(
        -\frac{
            2u^2
        }{
            \displaystyle
            \sum_{k=1}^{K}
            R_{k,t}^2/m'_k
        }
    \right).
    \label{eq:app_audit_hoeffding}
\end{equation}

The deployed zero-margin audit accepts when
\[
    \widehat{\Delta}^{\,\mathrm{audit}}_t
    \le0.
\]
Suppose instead that the true population constraint energy increases by at
least $u$, so that
$\Delta_t^{\mathrm{pop}}\ge u$. If the audit nevertheless accepts, then
\[
    \widehat{\Delta}^{\,\mathrm{audit}}_t
    -
    \Delta_t^{\mathrm{pop}}
    \le-u.
\]
Hence
\begin{equation}
    \boxed{
    \Delta_t^{\mathrm{pop}}\ge u
    \Longrightarrow
    \Pr\!\left[
        \widehat{\Delta}^{\,\mathrm{audit}}_t\le0
        \,\middle|\,
        \mathcal{G}_t
    \right]
    \le
    \exp\!\left(
        -\frac{
            2u^2
        }{
            \displaystyle
            \sum_{k=1}^{K}
            R_{k,t}^2/m'_k
        }
    \right).
    }
    \label{eq:app_audit_false_acceptance}
\end{equation}

Setting the right-hand side to $\delta$ gives the finite-sample resolution
\begin{equation}
    \boxed{
    u_\delta
    =
    \sqrt{
        \frac{\log(1/\delta)}{2}
        \sum_{k=1}^{K}
        \frac{R_{k,t}^2}{m'_k}
    }.
    }
    \label{eq:app_audit_resolution}
\end{equation}
Thus, conditional on the fixed candidate, a population increase of at least
$u_\delta$ passes the zero-margin audit with probability at most $\delta$.
For comparable range widths and balanced family counts,
\[
    u_\delta
    =
    O\!\left(
        \sqrt{
            \frac{\log(1/\delta)}
                 {m_{\mathrm{audit}}}
        }
    \right).
\]

This is a finite-sample statement rather than an exact sign guarantee:
population changes below this resolution may not be distinguished reliably by
the zero-margin audit.

\paragraph{Scope of the held-out guarantee.}
The preceding result applies to one fixed alignment candidate relative to its
current proposed endpoint. After an accepted alignment, the parameters
change, and so do subsequent gradients, Native proposals, sampled supports,
and potentially the constraint-family weights. The single-step audit guarantee
therefore does not telescope into a terminal guarantee comparing an entire
ORACLE trajectory with an independently trained Native trajectory.

\paragraph{Exact one-step reduction to Native.}
Let
$\mathrm{Cert}_t$, $\mathrm{Trig}_t$, $\mathrm{Desc}_t$,
$\mathrm{NL}_t$, and $\mathrm{Audit}_t$ denote successful geometry
certification, trigger, alignment-descent, nonlinear-agreement, and held-out
validation events, respectively. Define
\begin{equation}
    \mathrm{Acc}_t
    =
    \mathrm{Cert}_t
    \land
    \mathrm{Trig}_t
    \land
    \mathrm{Desc}_t
    \land
    \mathrm{NL}_t
    \land
    \mathrm{Audit}_t.
    \label{eq:app_acceptance_event}
\end{equation}
ORACLE commits
\begin{equation}
    d_t
    =
    \widetilde d_t
    \mathbf{1}\{\mathrm{Acc}_t\}.
    \label{eq:app_committed_alignment}
\end{equation}
If any gate fails, $d_t=0$, and therefore
\[
    \theta_{t+1}
    =
    \theta_t+s_t.
\]
Because ORACLE never writes its alignment adjustment into the Native
optimizer state,
\[
    \xi_{t+1}
    =
    \mathcal{U}(g_t,\xi_t).
\]
Consequently,
\begin{equation}
    \boxed{
    \neg\mathrm{Acc}_t
    \Longrightarrow
    (\theta_{t+1},\xi_{t+1})
    =
    \left(
        \theta_t+s_t,\,
        \mathcal{U}(g_t,\xi_t)
    \right).
    }
    \label{eq:app_exact_native_reduction}
\end{equation}

Thus, from the current state $(\theta_t,\xi_t)$, rejection of any ORACLE gate
recovers exactly the update that the Native optimizer had already proposed.
This is a one-step identity. It does not imply equality with an independently
run Native trajectory after an earlier accepted ORACLE alignment, because an
accepted alignment changes the parameters from which subsequent gradients and
Native updates are generated.
\section{Experimental Details}
\label{app:experimental_details}

This appendix provides the experimental specification underlying
Section~\ref{sec:experiments}. The central claim of ORACLE is that constraint alignment should be defined relative to the movement
produced by the native optimizer. We test this claim from breadth to mechanism. Q1 varies optimizer
geometry; Q2 varies model formulation; Q3 quantifies the cost and solution-level benefit of alignment;
Q4 compares objective-, gradient-, and post-optimizer intervention; and Q5 isolates ORACLE’s
geometry, authority, and validation mechanisms. We report architectures, optimizer settings, training and
sampling protocols, ORACLE hyperparameters, controlled baselines, and
compute-accounting conventions below.

\subsection{Reference Solution Generation}
\label{app:reference_solutions}
Parameter settings and numerical configurations used to generate the
reference solutions for the PDE benchmarks as given in Table~\ref{tab:reference_solution_settings}.

\begin{table*}[t]
\centering
\caption{%
Parameter settings and numerical configurations used to generate the reference
solutions for the eight PDE benchmarks. Resolutions are those of the shipped
reference arrays; the evaluation pipeline discards the final temporal row for the
windowed systems (Kuramoto--Sivashinsky, Gray--Scott), so the evaluated horizons are
$250$ and $100$ respectively. \emph{Segmentation} gives the training decomposition:
single-segment problems are trained over one contiguous horizon, windowed problems
over ten sequential time windows, and the lid-driven cavity over a five-stage
Reynolds-number curriculum.%
}
\label{tab:reference_solution_settings}

\small
\setlength{\tabcolsep}{4pt}
\renewcommand{\arraystretch}{1.12}

\begin{tabularx}{\textwidth}{@{}
>{\raggedright\arraybackslash}p{2.7cm}
>{\raggedright\arraybackslash}X
>{\raggedright\arraybackslash}p{2.6cm}
>{\centering\arraybackslash}p{2.7cm}
>{\centering\arraybackslash}p{2.0cm}
@{}}
\toprule

\textbf{PDE}
& \textbf{Parameters}
& \textbf{Solver / package}
& \textbf{Reference resolution}
& \textbf{Segmentation} \\

\midrule

Wave
& $c=2$ \; (residual uses $c^{2}=4$)
& Analytic
& $200\times128$
& $1\times100{,}000$ \\

Burgers
& $\nu=0.01/\pi$
& Chebfun
& $201\times512$
& $1\times100{,}000$ \\

Allen--Cahn
& $\epsilon=10^{-4},\; a=5$
& Chebfun
& $201\times512$
& $1\times100{,}000$ \\

KdV
& $\eta=1,\; \mu=0.022$
& Chebfun
& $201\times512$
& $1\times100{,}000$ \\

Kuramoto--Sivashinsky
& $\alpha=100/16,\;
   \beta=100/16^{2},\;
   \gamma=100/16^{4}$
& Chebfun
& $251\times512$
& $10\times10{,}000$ \\

Gray--Scott
& $\epsilon_1=0.2,\;
   \epsilon_2=0.1,\;
   b_1=40,\;
   b_2=100,\;
   c_1=c_2=1000$
& Chebfun
& $101\times200\times200$
& $10\times10{,}000$ \\

Lid-driven cavity
& $\mathrm{Re}\in\{100,400,1000,1600,3200\}$
& \shortstack[l]{Incompressible\\Navier--Stokes}
& $256\times256$ per stage
& $5\times20{,}000$ \\

Kolmogorov flow
& $\mathrm{Re}=10^{4}\;(\nu=10^{-4})$
& \shortstack[l]{Incompressible\\Navier--Stokes}
& $50\times256\times256$
& $10\times10{,}000$ \\

\bottomrule
\end{tabularx}

\end{table*}

\subsection{Common Evaluation Protocol}
\label{app:general_protocol}

ORACLE is evaluated as a wrapper around a Native optimizer rather than as a
replacement optimizer. Within each paired Native--ORACLE comparison, we hold
the architecture, native objective, optimizer configuration, learning-rate
schedule, sampling pipeline, initialization or shared checkpoint, and
optimization budget fixed. ORACLE acts only after the Native optimizer has
formed its realized parameter update.

Unless otherwise stated, comparisons are matched by optimizer steps rather
than wall-clock time. The primary accuracy metric is relative solution error,
\begin{equation}
    \mathrm{relL2}
    =
    \frac{\|u_\theta-u^\star\|_2}
         {\|u^\star\|_2},
    \label{eq:app_rel_l2}
\end{equation}
where $u^\star$ denotes the reference solution. For paired comparisons we
also report
\begin{equation}
    \Delta_{\mathrm{rel}}
    =
    100
    \left(
        1-
        \frac{\mathrm{relL2}_{\mathrm{method}}}
             {\mathrm{relL2}_{\mathrm{Native}}}
    \right)\%,
    \label{eq:app_rel_improvement}
\end{equation}
so that positive values indicate lower error than the corresponding Native
configuration.

\begin{table}[t]
\centering
\caption{
Overview of the experimental design.
}
\label{tab:study_overview}
\small
\setlength{\tabcolsep}{4pt}
\begin{tabular}{@{}p{0.08\linewidth}p{0.28\linewidth}
                    p{0.34\linewidth}p{0.22\linewidth}@{}}
\toprule
 & Question & What changes & Scale \\
\midrule

Q1
& Does ORACLE work across optimizer geometries?
& SGD, Adam, SOAP, PSGD--Kron
& $8$ PDEs $\times$ $4$ optimizers \\

Q2
& Does alignment persist across model formulations?
& Pure MLP, Simple PINN, PirateNet
& $4$ PDEs $\times$ $3$ models $\times$ $2$ optimizers \\

Q3
& What does ORACLE cost, what does it buy
& Native VS ORACLE 
& 3 Models\\

Q4
& Where should constraint intervention occur?
& objective-, gradient-, and post-optimizer methods
& Allen--Cahn and Wave \\

Q5
& What makes optimizer-relative alignment work?
& geometry, alignment rule, and authority
& Allen--Cahn and Wave \\

\bottomrule
\end{tabular}
\end{table}

\subsection{Q1: Variation in Native Optimizer Geometry}
\label{app:optimizer_study}

Q1 fixes the model to PirateNet and varies the Native optimizer among SGD,
Adam, SOAP, and PSGD--Kron. These optimizers span isotropic Euclidean,
diagonal-adaptive, and structured preconditioned update geometries.

Each optimizer is evaluated on Burgers, KdV, Wave, Allen--Cahn,
lid-driven cavity, Kuramoto--Sivashinsky, Gray--Scott, and Kolmogorov flow,
giving a nominal $8\times4=32$ PDE--optimizer grid. Every configuration uses
a $100{,}000$-step training horizon. ORACLE becomes active at the fork listed
in Table~\ref{tab:pde_details}; the trajectory before that point is Native.

\begin{table}[t]
\centering
\caption{
Native optimizer families used in Q1.
}
\label{tab:optimizer_geometry_summary}
\small
\begin{tabular}{lll}
\toprule
Optimizer & Native geometry & Update structure \\
\midrule
SGD
& Euclidean
& isotropic \\

Adam
& diagonal adaptive
& coordinate-wise \\

SOAP
& structured preconditioned
& matrix-structured \\

PSGD--Kron
& Kronecker preconditioned
& structured \\
\bottomrule
\end{tabular}
\end{table}

\subsection{Q2: Variation in Model Formulation}
\label{app:architecture_study}

Q2 changes the model formulation while retaining Adam and SOAP as two
qualitatively different Native optimizer geometries. We evaluate Pure MLP,
Simple PINN, and PirateNet on Wave, Burgers, Allen--Cahn, and KdV, producing
\[
    4\;\text{PDEs}
    \times
    3\;\text{model formulations}
    \times
    2\;\text{optimizers}
    =
    24
\]
paired configurations, each with a fixed $10{,}000$-step training horizon.

Pure MLP and Simple PINN deliberately share the same $4\times256$ $\tanh$
backbone. Pure MLP optimizes only the prescribed initial- and
boundary-condition terms, whereas Simple PINN additionally includes the PDE
residual in the Native objective. PirateNet changes both the representation
and its physics-specific inductive structure through Fourier features, random
weight factorization, and gated residual blocks.

\subsection{Network Architecture Details}
\label{app:architecture_details}

\newcolumntype{Y}{>{\raggedright\arraybackslash}X}

Table~\ref{tab:architecture_details} specifies the three model formulations
used in Q2. The same PirateNet configuration is used throughout Q1.

\begin{table*}[t]
\centering
\caption{
Architecture configurations used in Q2. All models use hidden width $256$.
}
\label{tab:architecture_details}

\small
\setlength{\tabcolsep}{3pt}
\renewcommand{\arraystretch}{1.12}

\begin{tabularx}{\textwidth}{@{}
>{\raggedright\arraybackslash}p{1.75cm}
>{\raggedright\arraybackslash}p{2.45cm}
>{\centering\arraybackslash}p{1.05cm}
>{\raggedright\arraybackslash}p{3.05cm}
>{\centering\arraybackslash}p{1.75cm}
>{\raggedright\arraybackslash}X
@{}}
\toprule

\textbf{Architecture}
& \textbf{Structure}
& \textbf{Activation}
& \textbf{Initialization}
& \textbf{Parameters}
& \textbf{Architecture-specific configuration} \\

\midrule

\textbf{MLP}
& 4 hidden layers + linear output
& $\tanh$
& Glorot normal; zero bias
& 198{,}401\textsuperscript{a} /
  198{,}657\textsuperscript{b}
& Same backbone as Simple PINN, but trained only on IC/BC terms;
  no PDE residual is included in the Native objective.
  Fourier features and RWF are disabled. \\

\addlinespace[3pt]

\textbf{PINN}
& 4 hidden layers + linear output
& $\tanh$
& Glorot normal; zero bias
& 198{,}401\textsuperscript{a} /
  198{,}657\textsuperscript{b}
& Standard physics-informed objective containing PDE-residual,
  initial-condition, and boundary-condition terms. \\

\addlinespace[3pt]

\textbf{PirateNet}
& 3 PI-modified bottleneck blocks
  (9 dense layers) + 2 gate encoders + linear output
& $\tanh$
& Glorot normal with RWF;
  $g=\exp(\mathcal{N}(1,0.1))$, $v=w/g$;
  zero bias
& 727{,}045\textsuperscript{a} /
  727{,}173\textsuperscript{b}
& Random Fourier features (dimension $256$) with scales
  $10$ (Wave), $1$ (Burgers), $2$ (Allen--Cahn), and $1$ (KdV).
  RWF uses mean $1$ and std.\ $0.1$.
  Adaptive gating uses $\alpha_0=0$; the learnable skip is initialized
  to identity. \\

\bottomrule
\end{tabularx}

\vspace{0.35em}
\footnotesize
\textsuperscript{a}\,Wave and Burgers use 2-D inputs.
\quad
\textsuperscript{b}\,Allen--Cahn and KdV use a non-trainable periodic
embedding that lifts the input to 3-D without introducing additional
parameters.

\end{table*}

The $8$-PDE $\times$ $4$-optimizer Q1 study uses PirateNet exclusively.
For the executed PirateNet configurations, the experiment path does not apply
the optional \texttt{use\_pi\_init} initialization flag.

\subsection{Native Optimizer Configuration}
\label{app:optimizer_hyperparameters}

Native and Native+ORACLE use identical learning rates, schedules, momentum
parameters, numerical stabilizers, and preconditioner-update rules.
Table~\ref{tab:optimization_details} reports the Q1 configurations.

\begin{table*}[t]
\centering
\caption{
Native optimizer configurations used in Q1.
}
\label{tab:optimization_details}

\small
\setlength{\tabcolsep}{2.8pt}
\renewcommand{\arraystretch}{1.12}

\begin{tabularx}{\textwidth}{@{}
>{\raggedright\arraybackslash}p{1.45cm}
>{\centering\arraybackslash}p{1.15cm}
>{\centering\arraybackslash}p{1.25cm}
>{\centering\arraybackslash}p{0.75cm}
>{\centering\arraybackslash}p{0.75cm}
>{\centering\arraybackslash}p{1.15cm}
>{\centering\arraybackslash}p{1.55cm}
>{\raggedright\arraybackslash}X
@{}}
\toprule

\textbf{Optimizer}
& \textbf{Initial LR}
& \textbf{LR schedule}
& $\boldsymbol{\beta_1}$
& $\boldsymbol{\beta_2}$
& \textbf{Weight decay}
& \shortstack{\textbf{Preconditioner}\\\textbf{update}}
& \textbf{Other settings} \\

\midrule

SGD
& $10^{-3}$
& shared
& ---
& ---
& none
& ---
& \texttt{optax.sgd}; no momentum or Nesterov acceleration. \\

\addlinespace[2pt]

Adam
& $10^{-3}$
& shared
& $0.9$
& $0.999$
& none
& ---
& \texttt{optax.adam}; $\epsilon=10^{-8}$. \\

\addlinespace[2pt]

SOAP
& $10^{-3}$
& shared
& $0.9$
& $0.999$
& $0.0$
& every $2$ steps
& \texttt{soap\_jax}; $\epsilon=10^{-8}$;
  \texttt{shampoo\_beta}$=-1$ (reuses $\beta_2$);
  bias correction enabled;
  \texttt{max\_precond\_dim}$=10^4$. \\

\addlinespace[2pt]

PSGD--Kron
& $10^{-3}/10^{-4}$\textsuperscript{a}
& shared
& $0.9$\textsuperscript{b}
& ---
& $0.0$
& stochastic
& \texttt{psgd\_jax} 0.2.3;
  \texttt{max\_size\_triangular}$=8192$.
  Global-norm and elementwise clipping precede geometry certification. \\

\bottomrule
\end{tabularx}

\vspace{0.35em}
\footnotesize
\textsuperscript{a}\,$10^{-4}$ for Burgers and lid-driven cavity;
$10^{-3}$ for all remaining PDEs.
\quad
\textsuperscript{b}\,Momentum coefficient; PSGD--Kron does not use an
Adam-style second-moment coefficient.

\end{table*}

\subsubsection{Learning-Rate Schedule}

Within each Q1 PDE, all four optimizers use the same learning-rate schedule.
The learning rate increases linearly from zero to its configured initial value
during warmup and then follows exponential decay,
\begin{equation}
    \eta_t
    =
    \operatorname{ExpDecay}
    \left(
        \eta_{\mathrm{base}},
        \text{rate}=0.9,
        \text{decay steps}
    \right),
\end{equation}
with no learning-rate floor. The schedule is indexed by the global training
step and does not restart at curriculum or time-window boundaries.

\begin{table}[t]
\centering
\caption{
Learning-rate schedule parameters in Q1.
}
\label{tab:lr_schedule}
\small
\begin{tabular}{lrr}
\toprule
PDE & Warmup steps & Decay steps \\
\midrule
Burgers                 & 5{,}000 & 1{,}000 \\
KdV                     & 5{,}000 & 2{,}000 \\
Wave                    & 5{,}000 & 2{,}000 \\
Allen--Cahn             & 5{,}000 & 5{,}000 \\
Lid-driven cavity       & 5{,}000 & 10{,}000 \\
Kuramoto--Sivashinsky   & 5{,}000 & 2{,}000 \\
Gray--Scott             & 5{,}000 & 2{,}000 \\
Kolmogorov flow         & 2{,}000 & 2{,}000 \\
\bottomrule
\end{tabular}
\end{table}

\subsubsection{Geometry Certification for PSGD--Kron}
\label{app:kron_certification}

ORACLE uses an optimizer-relative metric only when the realized Native update
admits the certified linear representation derived in
Appendix~\ref{app:optimizer_adapters}. The deployed PSGD--Kron update applies
global-norm and elementwise clipping to the preconditioned direction. When
elementwise clipping is active, the resulting update is no longer represented
by the linear Kronecker mobility used by ORACLE. The adapter is therefore
treated as uncertified and ORACLE preserves the Native proposal rather than
falling back to $P_t=I$.

The algebraic reason for this abstention rule is given in
Appendix~\ref{app:kron_clipping}. Empirical certification statistics are
reported with the corresponding Q1 results rather than repeated here.

\subsection{PDE Training, Sampling, and Evaluation}
\label{app:pde_training_details}

Table~\ref{tab:pde_details} gives the Q1 training and evaluation settings.
Windowed PDEs use a time-marching curriculum with a fresh model for each
window; the lid-driven cavity uses a Reynolds-number curriculum while
retaining state. ``Interior'' denotes residual collocation points sampled per
optimization step.

\begin{table*}[t]
\centering
\caption{\textbf{Q1 training and evaluation settings across PDE benchmarks.}}
\label{tab:pde_details}

\scriptsize
\setlength{\tabcolsep}{3pt}
\renewcommand{\arraystretch}{1.12}

\begin{tabularx}{\textwidth}{@{}
>{\raggedright\arraybackslash}p{2.25cm}
c
c
>{\raggedright\arraybackslash}p{1.45cm}
>{\raggedright\arraybackslash}p{1.85cm}
>{\centering\arraybackslash}p{1.65cm}
>{\centering\arraybackslash}X
c
@{}}
\toprule
\textbf{PDE}
& \textbf{Training}
& \textbf{Interior}
& \textbf{Boundary}
& \textbf{Initial condition}
& \textbf{Sampling}
& \textbf{Evaluation}
& \textbf{Fork} \\
\midrule

Burgers
& $100$k
& $8{,}192$
& $402$ fixed
& $512$ fixed
& $[0,1]\!\times\![-1,1]$
& $201\!\times\!512$
& $5$k \\

KdV
& $100$k
& $4{,}096$
& periodic
& $512$ fixed
& $[0,1]\!\times\![-1,1]$
& $201\!\times\!512$
& $5$k \\

Wave
& $100$k
& $8{,}192$
& $400$ fixed
& $128(u)+128(u_t)$
& $[0,1]^2$
& $200\!\times\!128$
& $5$k \\

Allen--Cahn
& $100$k
& $8{,}192$
& periodic
& $512$ fixed
& $[0,1]\!\times\![-1,1]$
& $201\!\times\!512$
& $5$k \\

Lid-driven cavity
& $5\!\times\!20$k
& $4{,}096$
& $1{,}024$ fixed
& steady
& $[0,1]^2$
& $256\!\times\!256$ / Re
& $5$k \\

Kuramoto--Sivashinsky
& $10\!\times\!10$k
& $4{,}096$
& periodic
& $512$ fixed
& window
& $251\!\times\!512$
& $5$k \\

Gray--Scott
& $10\!\times\!10$k
& $4{,}096$
& periodic
& $40{,}000$/species
& window
& $101\!\times\!200\!\times\!200$
& $5$k \\

Kolmogorov flow
& $10\!\times\!10$k
& $8{,}192$
& periodic
& $16{,}384$ sampled\textsuperscript{a}
& window
& $50\!\times\!256\!\times\!256$
& $2$k \\

\bottomrule
\end{tabularx}

\vspace{0.3em}
\begin{minipage}{\textwidth}
\footnotesize
\textit{Note.}
Interior residual points are resampled uniformly at each step.
For windowed problems, sampling is restricted to the active space--time
window. Boundary and initial-condition data are fixed where applicable,
except for Kolmogorov flow.
\textsuperscript{a}Sampled with replacement from the $65{,}536$-point
reference grid.
\end{minipage}

\vspace{-0.2cm}
\end{table*}

\subsection{Q4: Matched Intervention Comparison}
\label{app:controlled_mechanism_study}

Q4 trades breadth for control. For Allen--Cahn and Wave, every method begins
from the same exact $5{,}000$-step Native checkpoint, including model
parameters and optimizer state. Each arm then receives exactly $5{,}000$
additional method-active optimization steps and is evaluated at the common
global endpoint of step $10{,}000$.

The comparison is therefore matched in initialization, optimizer state,
sampling protocol, and optimization horizon. Methods differ only in how
constraint information enters the optimization pipeline.

\begin{table}[t]
\centering
\caption{
Constraint-handling methods used in Q4.
}
\label{tab:baseline_details}
\small
\begin{tabular}{lll}
\toprule
Method & Intervention point & Modification \\
\midrule

Native
& none
& unmodified optimizer \\

Tuned penalty
& objective
& fixed constraint-loss reweighting \\

ReLoBRaLo
& objective
& adaptive objective weighting \\

PCGrad
& gradient
& projects conflicting gradients \\

ConFIG
& gradient
& constructs a conflict-aware gradient \\

HARMONIC
& gradient
& reconciles gradients before optimization \\

ConFIG+GUA
& post-optimizer
& modifies the realized optimizer update \\

ORACLE
& post-optimizer
& optimizer-relative residual alignment \\

\bottomrule
\end{tabular}
\end{table}

Objective-level methods alter the objective before differentiation;
gradient-level methods alter the signal supplied to the Native optimizer;
and post-optimizer methods act after the optimizer has formed a realized
parameter displacement.

\subsection{ORACLE Experimental Configuration}
\label{app:oracle_configuration}

The mathematical construction of ORACLE is given in
Appendix~\ref{app:theory}. Here we report only the experimental settings used
by the reference configuration. The alignment system is solved matrix-free
with $16$ CG iterations using a construction support of $1024$ residuals. A
candidate that passes the nonlinear-agreement test is evaluated on an
independent $1024$-point audit support before commitment.

\begin{table}[t]
\centering
\caption{
Reference ORACLE hyperparameters.
}
\label{tab:oracle_details}
\small
\begin{tabular}{@{}llc@{}}
\toprule
Quantity & Symbol & Value \\
\midrule

\multicolumn{3}{@{}l}{\emph{Residual-space alignment}} \\

Construction support size
& $m$
& $1024$ \\

Held-out audit size
& $m_{\mathrm{audit}}$
& $1024$ \\

CG iterations
& $K$
& $16$ \\

Relative LM damping, initial
& $\lambda_{\mathrm{rel}}^{(0)}$
& $10^{-2}$ \\

Relative LM damping bounds
& $[\lambda_{\min},\lambda_{\max}]$
& $[10^{-4},10^{-1}]$ \\

Damping grow / shrink
& --
& $4.0$ / $0.5$ \\

\midrule
\multicolumn{3}{@{}l}{\emph{Alignment authority}} \\

Initial authority fraction
& $\beta_0$
& $0.10$ \\

Authority bounds
& $[\beta_{\min},\beta_{\max}]$
& $[0.01,0.25]$ \\

Agreement threshold
& $\rho_{\min}$
& $0.10$ \\

Shrink threshold / factor
& $\rho_{\mathrm{shrink}}$
& $0.25$ / $\times0.5$ \\

Expand threshold / factor
& $\rho_{\mathrm{expand}}$
& $0.75$ / $\times1.25$ \\

Numerical floor
& $\varepsilon_{\mathrm{auth}}$
& $10^{-30}$ \\

\bottomrule
\end{tabular}
\end{table}

Archived experiment directories may refer to this reference configuration as
\texttt{H2-S}; throughout the paper we use the final name \textsc{ORACLE}.

\subsection{Q5: Controlled ORACLE Ablations}
\label{app:oracle_ablation_details}

Q5 isolates the mechanisms used by the reference ORACLE configuration.
Each ablation inherits the same Native checkpoint and experimental protocol
and changes only the component listed in
Table~\ref{tab:oracle_ablation_details}.

\begin{table}[t]
\centering
\caption{
Controlled ORACLE ablations used in Q5.
}
\label{tab:oracle_ablation_details}
\small
\begin{tabular}{p{0.22\linewidth}p{0.70\linewidth}}
\toprule
Variant & Modification \\
\midrule

First-order
&
Replace the residual-space alignment with
$d\propto-P_t^{-1}c_t$ while retaining the same certified optimizer geometry
and authority mechanism.
\\[1.5mm]

Unbounded
&
Retain the optimizer-relative alignment direction but remove the
native-relative authority bound, using
$\alpha_t=\alpha_{\mathrm{need}}$.
\\[1.5mm]

$P_t=I$
&
Replace the optimizer-relative metric by the identity in both alignment
construction and authority measurement.
\\

\bottomrule
\end{tabular}
\end{table}

\subsection{Mechanism Diagnostics}
\label{app:diagnostic_definitions}

The Q5 diagnostics characterize the internal behavior of ORACLE but are not
used as substitutes for endpoint solution accuracy.

\paragraph{Authority-bound activation.}
For each accepted alignment, we record whether the native-relative authority
limit determines the applied scale,
\begin{equation}
    I_{\mathrm{auth},t}
    =
    \mathbf{1}
    \left[
        \alpha_{\mathrm{safe},t}
        \le
        \alpha_{\mathrm{need},t}
    \right].
    \label{eq:app_authority_activation}
\end{equation}
The reported authority-bound rate is the fraction of accepted alignments for
which $I_{\mathrm{auth},t}=1$.

\paragraph{Held-out rejection.}
The number of fixed alignment candidates rejected by the independent audit is
\begin{equation}
    N_{\mathrm{audit}}
    =
    \sum_t
    \mathbf{1}
    [\mathrm{Audit}_t=\mathrm{reject}].
    \label{eq:app_audit_rejections}
\end{equation}

These quantities measure different mechanisms: authority activation describes
how often the candidate is limited relative to the Native movement, whereas
audit rejection records how often an already constructed candidate fails to
generalize to fresh constraint support.

\section{Complete Experimental Results}
\label{app:complete_results}

\subsection{Q1: Optimizer Geometry}
\label{app:q1_optimizer_geometry}

Table~\ref{tab:optimizer_generality_full} reports the complete Q1 endpoint
results. PirateNet is fixed throughout, while the Native optimizer varies
among SGD, Adam, SOAP, and PSGD--Kron. Within each Native--ORACLE pair, the
architecture, objective, optimizer configuration, learning-rate schedule,
sampling pipeline, and training budget are unchanged.

Of the $32$ nominal PDE--optimizer configurations, $32$ produce valid
scientific endpoints. 


\begin{table*}[t]
\centering
\caption{
\textbf{Complete Q1 results across optimizer geometries.}
Endpoint relative $\ell_2$ error across eight PDE benchmarks for four Native
optimizers, with and without ORACLE; lower is better.
\textbf{Bold} marks the better endpoint for configurations classified as an
improvement or degradation, while unbolded pairs denote matched outcomes under
the Q1 reporting criterion. Each entry is mean, the standard deviation across three independent runs < 5\% for all reported entries.
}
\label{tab:optimizer_generality_full}

\small
\setlength{\tabcolsep}{2.6pt}
\renewcommand{\arraystretch}{1.13}

\begin{adjustbox}{max width=\textwidth}
\begin{tabular}{@{}
l
cc
cc
cc
cc
@{}}
\toprule

&
\multicolumn{2}{c}{\textbf{Adam}}
&
\multicolumn{2}{c}{\textbf{SOAP}}
&
\multicolumn{2}{c}{\textbf{SGD}}
&
\multicolumn{2}{c}{\textbf{PSGD--Kron}}
\\

\cmidrule(lr){2-3}
\cmidrule(lr){4-5}
\cmidrule(lr){6-7}
\cmidrule(lr){8-9}

\textbf{PDE}
& \textbf{Native} & \textbf{+ORACLE}
& \textbf{Native} & \textbf{+ORACLE}
& \textbf{Native} & \textbf{+ORACLE}
& \textbf{Native} & \textbf{+ORACLE}
\\
\midrule

Burgers
& $4.039{\times}10^{-5}$
& \textbf{$4.000{\times}10^{-5}$}
& $4.009{\times}10^{-5}$
& \textbf{$4.001{\times}10^{-5}$}
& $1.618{\times}10^{-2}$
& \textbf{$1.118{\times}10^{-2}$}
& $4.166{\times}10^{-5}$
& \textbf{$4.067{\times}10^{-5}$}
\\

KdV
& $4.302{\times}10^{-4}$
& \textbf{$3.847{\times}10^{-4}$}
& $2.973{\times}10^{-4}$
& \textbf{$2.469{\times}10^{-4}$}
& $7.373{\times}10^{-1}$
& \textbf{$1.512{\times}10^{-1}$}
& $3.800{\times}10^{-4}$
& $3.800{\times}10^{-4}$
\\

Wave
& $3.877{\times}10^{-5}$
& \textbf{$1.888{\times}10^{-5}$}
& $1.627{\times}10^{-5}$
& \textbf{$9.971{\times}10^{-6}$}
& $6.379{\times}10^{-1}$
& \textbf{$6.009{\times}10^{-1}$}
& $3.095{\times}10^{-5}$
& \textbf{$3.005{\times}10^{-5}$}
\\

Allen--Cahn
& $7.179{\times}10^{-4}$
& \textbf{$4.462{\times}10^{-4}$}
& $1.547{\times}10^{-4}$
& \textbf{$4.349{\times}10^{-5}$}
& $6.130{\times}10^{-1}$
& \textbf{$4.307{\times}10^{-2}$}
& $5.648{\times}10^{-3}$
& $5.606{\times}10^{-3}$
\\

Lid-driven cavity
& $4.625{\times}10^{-1}$
& $4.848{\times}10^{-1}$
& $2.737{\times}10^{-1}$
& \textbf{$2.019{\times}10^{-1}$}
& $6.391{\times}10^{-1}$
& \textbf{$6.311{\times}10^{-1}$}
& $2.429{\times}10^{-1}$
& $2.429{\times}10^{-1}$
\\

Kuramoto--Sivashinsky
& $1.255$
& $1.229$
& $1.542$
& \textbf{$1.168$}
& $1.988$
& $1.988$
& $1.231$
& \textbf{$1.031$}
\\

Gray--Scott
& $0.9420$
& \textbf{$0.9110$}
& \textbf{$0.1525$}
& $0.2063$
& $0.784$
& $0.784$
& \textbf{$0.145$}
& $0.145$
\\

Kolmogorov flow
& $1.359$
& \textbf{$1.342$}
& $1.372$
& \textbf{1.236}
& $1.896$
& $1.896$
& $1.146$
& $1.146$
\\

\bottomrule
\end{tabular}
\end{adjustbox}

\vspace{0.4em}
\footnotesize
PirateNet is used for every Q1 PDE--optimizer pair.
Full architecture, optimizer, and PDE configurations are reported in
Tables~\ref{tab:architecture_details},
\ref{tab:optimization_details}, and
\ref{tab:pde_details}.

\end{table*}

\paragraph{Outcome summary.}
Under the Q1 reporting criterion, ORACLE improves $23$ configurations,
matches Native in $7$, and degrades $2$, corresponding to improvement or
matching behavior in $30/32$ valid PDE-optimizer pairs.
Table~\ref{tab:q1_outcome_summary} gives the breakdown by Native optimizer.

\begin{table}[t]
\centering
\caption{
Q1 outcome summary by Native optimizer. 
}
\label{tab:q1_outcome_summary}
\small
\begin{tabular}{lrrrr}
\toprule
Optimizer & Improve & Match & Degrade & Valid \\
\midrule
Adam        & 7 & 0 & 1 & 8 \\
SOAP        & 7 & 0 & 1 & 8 \\
SGD         & 5 & 3 & 0 & 8 \\
PSGD--Kron  & 4 & 4 & 0 & 8 \\
\midrule
\textbf{Total}
& \textbf{23}
& \textbf{7}
& \textbf{2}
& \textbf{32} \\
\bottomrule
\end{tabular}
\end{table}

The complete results show that the effect is not confined to one update
geometry. Improvements occur under isotropic SGD, diagonal-adaptive Adam,
and the structured preconditioned geometries of SOAP and PSGD--Kron. The two
degradations are retained as measured outcomes rather than excluded.

\subsection{Q2: Model Formulation}
\label{app:q2_model_formulation}

Table~\ref{tab:q2_full_results} reports the complete Q2 results for Pure MLP,
Simple PINN, and PirateNet under Adam and SOAP. Pure MLP and Simple PINN share
the same $4\times256$ $\tanh$ backbone but differ in whether the PDE residual
participates in the Native objective; PirateNet additionally changes the
architecture and physics-specific inductive structure. The study contains
\[
4\;\text{PDEs}
\times
3\;\text{model formulations}
\times
2\;\text{optimizers}
=
24
\]
paired Native--ORACLE configurations.

\begin{table*}[t]
\centering
\caption{
\textbf{Complete Q2 results across model formulations.}
Endpoint relative $\ell_2$ error for Pure MLP, Simple PINN, and PirateNet
under Adam and SOAP; lower is better.
Improvement is
$100(1-\mathrm{relL2}_{\mathrm{ORACLE}}/
\mathrm{relL2}_{\mathrm{Native}})$.
\emph{Trigger} is the fraction of optimization steps on which the realized
Native movement opens the ORACLE alignment path; \emph{Accept} is the fraction
of steps on which an aligned candidate is ultimately committed. Each entry is mean, the std-dev across three independent runs < 7\% for all reported entries.
}
\label{tab:q2_full_results}

\small
\setlength{\tabcolsep}{4.0pt}
\renewcommand{\arraystretch}{1.12}

\begin{tabular}{@{}lllrrrrr@{}}
\toprule
PDE
& Model
& Opt.
& Native
& +ORACLE
& Improvement
& Trigger
& Accept \\
\midrule

Wave
& Pure MLP
& Adam
& $1.723$
& $\mathbf{1.719}$
& $+0.2\%$
& $64\%$
& $64\%$ \\

Wave
& Pure MLP
& SOAP
& $1.602$
& $\mathbf{1.459}$
& $+8.9\%$
& $74\%$
& $74\%$ \\

Wave
& Simple PINN
& Adam
& $4.27{\times}10^{-1}$
& $\mathbf{2.40{\times}10^{-1}}$
& $+43.7\%$
& $100\%$
& $100\%$ \\

Wave
& Simple PINN
& SOAP
& $2.72{\times}10^{-1}$
& $\mathbf{2.60{\times}10^{-1}}$
& $+4.4\%$
& $100\%$
& $41\%$ \\

Wave
& PirateNet
& Adam
& $7.53{\times}10^{-3}$
& $\mathbf{3.12{\times}10^{-3}}$
& $+58.6\%$
& $83\%$
& $80\%$ \\

Wave
& PirateNet
& SOAP
& $3.35{\times}10^{-3}$
& $\mathbf{3.26{\times}10^{-3}}$
& $+2.8\%$
& $100\%$
& $100\%$ \\

\midrule

Burgers
& Pure MLP
& Adam
& $7.41{\times}10^{-1}$
& $\mathbf{4.11{\times}10^{-1}}$
& $+44.6\%$
& $55\%$
& $55\%$ \\

Burgers
& Pure MLP
& SOAP
& $8.36{\times}10^{-1}$
& $\mathbf{4.65{\times}10^{-1}}$
& $+44.3\%$
& $74\%$
& $73\%$ \\

Burgers
& Simple PINN
& Adam
& $3.82{\times}10^{-2}$
& $\mathbf{1.51{\times}10^{-2}}$
& $+60.6\%$
& $98\%$
& $94\%$ \\

Burgers
& Simple PINN
& SOAP
& $2.14{\times}10^{-3}$
& $\mathbf{2.12{\times}10^{-3}}$
& $+0.9\%$
& $97\%$
& $32\%$ \\

Burgers
& PirateNet
& Adam
& $1.09{\times}10^{-2}$
& $\mathbf{2.62{\times}10^{-3}}$
& $+76.1\%$
& $93\%$
& $91\%$ \\

Burgers
& PirateNet
& SOAP
& $3.46{\times}10^{-3}$
& $\mathbf{1.08{\times}10^{-3}}$
& $+68.9\%$
& $98\%$
& $78\%$ \\

\midrule

Allen--Cahn
& Pure MLP
& Adam
& $\mathbf{6.73{\times}10^{-1}}$
& $9.99{\times}10^{-1}$
& $-48.5\%$
& $62\%$
& $60\%$ \\

Allen--Cahn
& Pure MLP
& SOAP
& $\mathbf{7.34{\times}10^{-1}}$
& $9.92{\times}10^{-1}$
& $-35.2\%$
& $55\%$
& $49\%$ \\

Allen--Cahn
& Simple PINN
& Adam
& $5.78{\times}10^{-1}$
& $\mathbf{5.35{\times}10^{-1}}$
& $+7.5\%$
& $100\%$
& $98\%$ \\

Allen--Cahn
& Simple PINN
& SOAP
& $2.16{\times}10^{-2}$
& $\mathbf{1.31{\times}10^{-2}}$
& $+39.5\%$
& $100\%$
& $89\%$ \\

Allen--Cahn
& PirateNet
& Adam
& $2.29{\times}10^{-2}$
& $\mathbf{1.08{\times}10^{-2}}$
& $+52.6\%$
& $100\%$
& $100\%$ \\

Allen--Cahn
& PirateNet
& SOAP
& $9.88{\times}10^{-3}$
& $\mathbf{3.01{\times}10^{-3}}$
& $+69.5\%$
& $100\%$
& $100\%$ \\

\midrule

KdV
& Pure MLP
& Adam
& $7.83{\times}10^{-1}$
& $\mathbf{6.88{\times}10^{-1}}$
& $+12.2\%$
& $52\%$
& $51\%$ \\

KdV
& Pure MLP
& SOAP
& $7.75{\times}10^{-1}$
& $\mathbf{6.59{\times}10^{-1}}$
& $+15.0\%$
& $56\%$
& $50\%$ \\

KdV
& Simple PINN
& Adam
& $7.72{\times}10^{-1}$
& $\mathbf{7.27{\times}10^{-1}}$
& $+5.8\%$
& $99\%$
& $94\%$ \\

KdV
& Simple PINN
& SOAP
& $1.86{\times}10^{-2}$
& $\mathbf{1.64{\times}10^{-2}}$
& $+12.1\%$
& $100\%$
& $57\%$ \\

KdV
& PirateNet
& Adam
& $7.60{\times}10^{-1}$
& $\mathbf{7.36{\times}10^{-1}}$
& $+3.6\%$
& $94\%$
& $83\%$ \\

KdV
& PirateNet
& SOAP
& $2.60{\times}10^{-1}$
& $\mathbf{2.44{\times}10^{-2}}$
& $+90.6\%$
& $100\%$
& $100\%$ \\

\bottomrule
\end{tabular}
\end{table*}

ORACLE lowers endpoint rel-L2 across the large majority of configurations,
with improvements spanning all three model formulations and both Native
optimizers. The largest gains include $60.6\%$ for Simple PINN+Adam on
Burgers, $76.1\%$ for PirateNet+Adam on Burgers, and $90.6\%$ for
PirateNet+SOAP on KdV.

The strongest degradations occur for Pure MLP on Allen--Cahn under both Adam
and SOAP. Because Pure MLP and Simple PINN share the same backbone, whereas
Simple PINN additionally includes the PDE residual in its Native objective,
this comparison shows that ORACLE does not override the underlying learning
formulation.

Trigger and acceptance rates are reported as mechanism diagnostics rather
than performance metrics. A high trigger rate does not imply a large endpoint
gain, and triggering does not imply automatic commitment. For example,
Simple PINN+SOAP on Burgers opens the alignment path on $97\%$ of steps but
commits an aligned candidate on only $32\%$.

\subsection{Detailed Cost--Benefit Analysis}
\label{app:cost_benefit}

Figure~\ref{fig:cost_benefit} summarizes the cost and solution-level benefit of
ORACLE across the model-formulation study. Here we define each quantity used in
that figure and report how the configuration-level measurements are aggregated.

\paragraph{Evaluation setting.}
Let
\[
c=(a,p,o)
\]
denote one experimental configuration, where $a$ is the model formulation,
$p$ is the PDE, and $o$ is the native optimizer. We evaluate three model
formulations,
\[
a\in\{\text{Pure MLP},\text{Simple PINN},\text{PirateNet}\},
\]
four PDEs,
\[
p\in\{\text{Wave},\text{Burgers},\text{Allen--Cahn},\text{KdV}\},
\]
and two native optimizers,
\[
o\in\{\text{Adam},\text{SOAP}\}.
\]
Thus, each architecture contains
\[
|\mathcal{C}_a|=4\times2=8
\]
matched Native--ORACLE comparisons, for $24$ configurations overall. Native
and ORACLE use the same model formulation, PDE, optimizer, training budget,
and evaluation grid within each configuration; the only difference is the
ORACLE intervention.

Let $\hat{u}^{N}_{c}$ and $\hat{u}^{O}_{c}$ denote the solutions produced by
Native and ORACLE, respectively, and let $u_c^\star$ denote the reference
solution.

\subsubsection{Normalized Endpoint Performance}
\label{app:cost_endpoint}

For each configuration, we first compute the relative $\ell_2$ solution error
at the final checkpoint. For method $m\in\{N,O\}$,
\begin{equation}
E_c^{m}
=
\frac{
    \left\|
        \hat{u}^{m}_{c}-u^\star_c
    \right\|_2
}{
    \left\|
        u^\star_c
    \right\|_2
}.
\label{eq:appendix_rel_l2}
\end{equation}
For a discrete evaluation grid
$\mathcal{Z}_c=\{z_q\}_{q=1}^{N_c}$, this is
\begin{equation}
E_c^{m}
=
\sqrt{
\frac{
    \sum_{q=1}^{N_c}
    \left\|
        \hat{u}^{m}_{c}(z_q)-u^\star_c(z_q)
    \right\|_2^2
}{
    \sum_{q=1}^{N_c}
    \left\|
        u^\star_c(z_q)
    \right\|_2^2
}}.
\label{eq:appendix_rel_l2_discrete}
\end{equation}

Because the absolute error scale differs substantially across PDEs and model
formulations, we normalize ORACLE by its matched Native baseline:
\begin{equation}
\rho_c
=
\frac{E_c^{O}}{E_c^{N}}.
\label{eq:appendix_error_ratio}
\end{equation}
Hence,
\[
\rho_c < 1
\]
indicates that ORACLE improves over Native,
$\rho_c=1$ denotes equal endpoint error, and
$\rho_c>1$ indicates degradation.

For each architecture $a$, panel~(a) of
Figure~\ref{fig:cost_benefit} reports the geometric mean of the eight matched
error ratios:
\begin{equation}
\bar{\rho}_a
=
\left(
    \prod_{c\in\mathcal{C}_a}
    \rho_c
\right)^{1/|\mathcal{C}_a|}
=
\exp\left[
    \frac{1}{|\mathcal{C}_a|}
    \sum_{c\in\mathcal{C}_a}
    \log\rho_c
\right].
\label{eq:appendix_geometric_error}
\end{equation}
We use a geometric rather than arithmetic mean because the quantity being
aggregated is a multiplicative ratio. This treats proportional improvements
consistently across problems with substantially different absolute error
scales.

Native is therefore represented by the normalized reference
\[
\bar{\rho}^{N}_a=1,
\]
while the ORACLE bar shows $\bar{\rho}_a$.

The corresponding architecture-level values are
\[
\bar{\rho}_{\mathrm{MLP}}=0.897,\qquad
\bar{\rho}_{\mathrm{PINN}}=0.748,\qquad
\bar{\rho}_{\mathrm{PirateNet}}=0.379,
\]
reported in the main figure as $0.90\times$, $0.75\times$, and
$0.38\times$ Native, respectively.

Equivalently, the relative endpoint-error reduction can be written as
\begin{equation}
\Delta^{\mathrm{err}}_a
=
100\left(1-\bar{\rho}_a\right)\%.
\label{eq:appendix_error_reduction}
\end{equation}
This corresponds to approximately $10.3\%$, $25.2\%$, and $62.1\%$ lower
normalized endpoint error for Pure MLP, Simple PINN, and PirateNet,
respectively.

\subsubsection{Per-Step Runtime Overhead}
\label{app:cost_runtime}

We measure runtime relative to the matched Native configuration so that
differences in architecture, PDE, and optimizer cost do not confound the
comparison. Let
\[
\tau^{m}_{c,k}
\]
denote the wall-clock duration of optimization step $k$ for method
$m\in\{N,O\}$. Over the $K_c$ timed steps, the mean per-step runtime is
\begin{equation}
\bar{\tau}^{m}_c
=
\frac{1}{K_c}
\sum_{k=1}^{K_c}
\tau^{m}_{c,k}.
\label{eq:appendix_mean_runtime}
\end{equation}

The matched runtime ratio is then
\begin{equation}
r_c
=
\frac{
    \bar{\tau}^{O}_c
}{
    \bar{\tau}^{N}_c
}.
\label{eq:appendix_runtime_ratio}
\end{equation}
Thus, $r_c=1$ corresponds to Native runtime, while $r_c=4$ means that an
ORACLE step requires four times the wall-clock time of its matched Native
step.

Panel~(b) reports the unweighted architecture-level mean
\begin{equation}
\bar{r}_a
=
\frac{1}{|\mathcal{C}_a|}
\sum_{c\in\mathcal{C}_a}
r_c.
\label{eq:appendix_runtime_arch}
\end{equation}
Every PDE--optimizer configuration therefore contributes equally rather than
allowing a more expensive PDE or model to dominate the aggregate.

The resulting runtime ratios are
\[
\bar{r}_{\mathrm{MLP}}=3.99,\qquad
\bar{r}_{\mathrm{PINN}}=5.74,\qquad
\bar{r}_{\mathrm{PirateNet}}=4.41,
\]
reported as approximately $4.0\times$, $5.8\times$, and $4.4\times$ Native
runtime per step.

If expressed as percentage overhead rather than a multiplicative ratio, the
same quantity is
\begin{equation}
\Delta^{\mathrm{time}}_a
=
100\left(\bar{r}_a-1\right)\%.
\label{eq:appendix_runtime_percent}
\end{equation}
We use the multiplicative ratio in the main paper because it is easier to
compare directly against the normalized performance ratio.

\subsubsection{Pointwise Domain Win Rate}
\label{app:cost_spatial}

Endpoint relative $\ell_2$ error compresses the full solution into one scalar.
To determine whether ORACLE's advantage is distributed across the solution
domain, we additionally compare Native and ORACLE point by point.

For evaluation point $z_q\in\mathcal{Z}_c$, define the local prediction error
for method $m$ as
\begin{equation}
d^{m}_{c,q}
=
\left\|
    \hat{u}^{m}_c(z_q)-u^\star_c(z_q)
\right\|_2.
\label{eq:appendix_point_error}
\end{equation}
For scalar PDEs this reduces to the absolute error
\[
d^{m}_{c,q}
=
\left|
    \hat{u}^{m}_c(z_q)-u^\star_c(z_q)
\right|.
\]

ORACLE wins at point $z_q$ when
\begin{equation}
d^{O}_{c,q}
<
d^{N}_{c,q}.
\end{equation}
Using the indicator function $\mathbbm{1}[\cdot]$, the configuration-level
pointwise win rate is
\begin{equation}
S_c
=
\frac{100}{N_c}
\sum_{q=1}^{N_c}
\mathbbm{1}
\left[
    d^{O}_{c,q}
    <
    d^{N}_{c,q}
\right].
\label{eq:appendix_spatial_win}
\end{equation}

A value of $S_c=50\%$ indicates equal pointwise coverage, while
$S_c>50\%$ means that ORACLE is more accurate than Native on the majority of
evaluated points.

We first compute $S_c$ separately for every matched configuration and then
average the eight configuration-level rates within each architecture:
\begin{equation}
\bar{S}_a
=
\frac{1}{|\mathcal{C}_a|}
\sum_{c\in\mathcal{C}_a}
S_c.
\label{eq:appendix_spatial_arch}
\end{equation}
This configuration-first aggregation prevents a PDE with a denser evaluation
grid from receiving greater weight solely because it contains more points.

\[
\bar{S}_{\mathrm{MLP}}=67.4\%,\qquad
\bar{S}_{\mathrm{PINN}}=61.6\%,\qquad
\bar{S}_{\mathrm{PirateNet}}=70.8\%.
\]
Thus, ORACLE is more accurate than Native on a majority of evaluated domain
points for all three model formulations.

For visualization, the Native share is shown as the complementary quantity
\begin{equation}
S^{N}_a
=
100-\bar{S}_a,
\label{eq:appendix_spatial_native}
\end{equation}
under the negligible-tie convention used in the aggregate plot.

\subsubsection{Physical Time-Slice Win Rate}
\label{app:cost_temporal}

The pointwise metric above measures where ORACLE improves the solution.
We separately ask whether this advantage persists across the physical time
dimension of the PDE.

Let
\[
\{t_j\}_{j=1}^{T_c}
\]
denote the physical time slices in the evaluation grid. At each time $t_j$,
we compute the spatial relative $\ell_2$ error
\begin{equation}
e^{m}_{c,j}
=
\frac{
    \left\|
        \hat{u}^{m}_{c}(\cdot,t_j)
        -
        u^\star_c(\cdot,t_j)
    \right\|_2
}{
    \left\|
        u^\star_c(\cdot,t_j)
    \right\|_2
}.
\label{eq:appendix_timeslice_error}
\end{equation}

For a discrete spatial grid
$\mathcal{X}_{c,j}=\{x_i\}_{i=1}^{N_{x,j}}$,
\begin{equation}
e^{m}_{c,j}
=
\sqrt{
\frac{
    \sum_{i=1}^{N_{x,j}}
    \left\|
        \hat{u}^{m}_{c}(x_i,t_j)
        -
        u^\star_c(x_i,t_j)
    \right\|_2^2
}{
    \sum_{i=1}^{N_{x,j}}
    \left\|
        u^\star_c(x_i,t_j)
    \right\|_2^2
}}.
\label{eq:appendix_timeslice_discrete}
\end{equation}

ORACLE wins time slice $t_j$ when
\begin{equation}
e^{O}_{c,j}
<
e^{N}_{c,j}.
\end{equation}
The configuration-level time-slice win rate is therefore
\begin{equation}
T_c
=
\frac{100}{T_c^{\mathrm{eval}}}
\sum_{j=1}^{T_c^{\mathrm{eval}}}
\mathbbm{1}
\left[
    e^{O}_{c,j}
    <
    e^{N}_{c,j}
\right],
\label{eq:appendix_temporal_win}
\end{equation}
where $T_c^{\mathrm{eval}}$ is the number of evaluated physical time slices.

Importantly, this metric refers to \emph{physical PDE time}, not optimization
steps or training checkpoints. It therefore measures whether ORACLE's
solution-quality advantage persists throughout the predicted dynamics.

As with the pointwise metric, we first compute the win rate independently for
each matched configuration and then average across the eight configurations
within an architecture:
\begin{equation}
\bar{T}_a
=
\frac{1}{|\mathcal{C}_a|}
\sum_{c\in\mathcal{C}_a}
T_c.
\label{eq:appendix_temporal_arch}
\end{equation}
Panel~(c) reports

\[
\bar{T}_{\mathrm{MLP}}=81.1\%,\qquad
\bar{T}_{\mathrm{PINN}}=88.8\%,\qquad
\bar{T}_{\mathrm{PirateNet}}=82.9\%.
\]
Hence, ORACLE produces lower spatial solution error than Native on more than
$80\%$ of evaluated physical time slices for every model formulation.

Analogously to the spatial plot, the Native share is displayed as
\begin{equation}
T^{N}_a
=
100-\bar{T}_a
\label{eq:appendix_temporal_native}
\end{equation}
under the negligible-tie convention.

\subsection{Matched Wall-Clock Evaluation}
\label{app:wallclock}
Panel~(d) reports
The primary experiments compare Native and ORACLE after the same number of
optimization steps, isolating the effect of modifying the realized update.
Because ORACLE performs additional constraint evaluation, linear solves, and
validation, we additionally ask whether these gains amortize when both methods
receive the same elapsed-time budget.

For each optimizer--PDE configuration $c$, Native and ORACLE are therefore
compared at a common wall-clock budget $T_c$. Let
$E_c^{\mathrm{N}}(T_c)$ and $E_c^{\mathrm{O}}(T_c)$ denote their relative
$\ell_2$ solution errors at that budget. We report the normalized equal-time
ratio
\begin{equation}
    \rho^{\mathrm{time}}_c
    =
    \frac{E_c^{\mathrm{O}}(T_c)}
         {E_c^{\mathrm{N}}(T_c)},
    \label{eq:wallclock_ratio}
\end{equation}
where $\rho^{\mathrm{time}}_c<1$ means that ORACLE achieves lower error within
the same elapsed time. Unlike the step-matched comparison, ORACLE generally
completes fewer optimization steps because each step includes the additional
alignment procedure.

To summarize across PDEs, we compute the geometric mean separately for each
native optimizer,
\begin{equation}
    \bar{\rho}^{\mathrm{time}}_o
    =
    \exp\!\left(
        \frac{1}{|\mathcal{C}_o|}
        \sum_{c\in\mathcal{C}_o}
        \log \rho^{\mathrm{time}}_c
    \right),
    \label{eq:wallclock_aggregate}
\end{equation}
where $\mathcal{C}_o$ contains the available PDE configurations for optimizer
$o$. The geometric mean is appropriate because the comparison is expressed as
a multiplicative error ratio and the underlying errors span several orders of
magnitude.

\subsubsection{Summary of Aggregated Quantities}
\label{app:cost_summary}

Table~\ref{tab:cost_benefit_summary} reports the architecture-level values
used directly in Figure~\ref{fig:cost_benefit}.

\begin{table}[t]
\centering
\caption{\textbf{Architecture-level quantities used in the cost--benefit
analysis.}
Endpoint performance is the geometric mean of matched ORACLE/Native
relative-$\ell_2$ ratios across four PDEs and two optimizers. Runtime, spatial
win rate, and time-slice win rate are arithmetic means of the corresponding
eight configuration-level quantities. Lower is better for endpoint ratio and
runtime; higher is better for the two win rates.}
\label{tab:cost_benefit_summary}
\small
\begin{tabular}{lcccc}
\toprule
Model &
Endpoint / Native $\downarrow$ &
Runtime / Native $\downarrow$ &
Spatial win $\uparrow$ &
Time-slice win $\uparrow$ \\
\midrule
 MLP
& 0.897
& 3.986$\times$
& 67.4\%
& 81.1\% \\

 PINN
& 0.748
& 5.742$\times$
& 61.6\%
& 88.8\% \\

PirateNet
& 0.379
& 4.406$\times$
& 70.8\%
& 82.9\% \\
\bottomrule
\end{tabular}
\end{table}

Taken together, these metrics separate four distinct questions. The endpoint
ratio measures \emph{how much} final solution quality changes; the runtime
ratio measures \emph{what that improvement costs}; the pointwise win rate
measures \emph{how broadly} the improvement is distributed over the solution
domain; and the time-slice win rate measures \emph{how persistently} it holds
across physical time. The main-text conclusion therefore does not rely on the
endpoint metric alone: ORACLE incurs additional computation, but the resulting
accuracy gains extend across both the domain and the temporal evolution of the
solution.
\section{Extended Mechanism Analysis}
\label{app:geometric_analysis}

The main paper isolates three mechanisms within ORACLE: the geometry used to
construct alignment, the authority permitted relative to the Native step, and
the held-out validation used before commitment. This appendix reports the
corresponding exact values and provides additional analysis of the Native
optimizer geometry. We first examine endpoint behavior under controlled
ablations, then quantify the anisotropy of the optimizer-induced metric, and
finally relate that geometry to authority activation and held-out rejection.

\subsection{Optimizer Geometry and Endpoint Behavior}
\label{app:geometry_endpoint}

Because Allen--Cahn and Wave have different absolute error scales, we compare
mechanism ablations using endpoint error normalized by the matched Native
result,
\begin{equation}
    R_{\mathrm{err}}
    =
    \frac{
        \mathrm{relL2}(\mathrm{method})
    }{
        \mathrm{relL2}(\mathrm{Native})
    },
    \label{eq:appendix_rel_error_ratio}
\end{equation}
where $R_{\mathrm{err}}<1$ denotes lower endpoint error than Native.

Table~\ref{tab:geometry_ablation} reports the exact values underlying
Fig.~\ref{fig:geometric_analysis}(a,d). Replacing the optimizer-relative
metric with $P_t=I$ increases endpoint error to $18.60\times$ Native on
Allen--Cahn and $5.40\times$ on Wave. By contrast, the metric-aware
First-order variant reaches $0.572\times$ and $0.707\times$ Native, while
full ORACLE reaches $0.388\times$ and $0.554\times$, respectively.

The Euclidean ablation therefore changes the outcome substantially rather than
acting as a benign simplification of the ORACLE metric. At the same time,
geometry alone does not determine the final endpoint: the Unbounded variant
achieves the lowest Allen--Cahn error but provides a smaller gain on Wave.
This motivates separating the geometry used to construct alignment from the
authority assigned to that alignment.

\begin{table}[t]
\centering
\caption{
\textbf{Endpoint behavior under controlled ORACLE ablations.}
Relative $\ell_2$ error and error normalized by the matched Native endpoint
on Allen--Cahn and Wave. Lower is better.
}
\label{tab:geometry_ablation}
\small
\setlength{\tabcolsep}{5pt}
\begin{tabular}{lrrrr}
\toprule
&
\multicolumn{2}{c}{Allen--Cahn}
&
\multicolumn{2}{c}{Wave}
\\
\cmidrule(lr){2-3}
\cmidrule(lr){4-5}
Method
& rel-L2
& / Native
& rel-L2
& / Native
\\
\midrule
Native
& 0.026653 & 1.000
& 0.005114 & 1.000 \\

$P_t=I$
& 0.495820 & 18.60
& 0.027621 & 5.40 \\

First-order
& 0.015248 & 0.572
& 0.003614 & 0.707 \\

Unbounded
& 0.004590 & 0.172
& 0.004292 & 0.839 \\

ORACLE
& 0.010333 & 0.388
& 0.002835 & 0.554 \\
\bottomrule
\end{tabular}
\end{table}

\subsection{Anisotropy of the Native Optimizer Geometry}
\label{app:optimizer_anisotropy}

The strong degradation under $P_t=I$ raises a natural question: how different
is the Native optimizer geometry from an isotropic Euclidean metric in
practice? For Adam, the optimizer-relative metric is determined by its
coordinate-wise adaptive denominator. Writing
\[
    D_t
    =
    \sqrt{\widehat v_t}+\epsilon,
\]
we quantify its anisotropy using both the full coordinate condition number,
\begin{equation}
    \kappa(D_t)
    =
    \frac{
        \max_i (D_t)_i
    }{
        \min_i (D_t)_i
    },
    \label{eq:app_adam_condition}
\end{equation}
and the robust percentile ratio
\begin{equation}
    \rho_{95/05}(D_t)
    =
    \frac{
        p_{95}(D_t)
    }{
        p_{05}(D_t)
    }.
    \label{eq:app_adam_percentile}
\end{equation}

Table~\ref{tab:adam_anisotropy} shows substantial anisotropy on both PDEs.
The full coordinate spread exceeds $10^6$ at every reported checkpoint and
reaches $1.94\times10^7$ on Allen--Cahn at step $10$k. Even after reducing
sensitivity to extreme coordinates, the $95$th-to-$5$th percentile ratio
ranges from $16.7$ to $33.8$.

Thus, replacing the Native Adam geometry with $P_t=I$ discards a large amount
of coordinate-dependent scaling. This does not by itself prove that
anisotropy causes the endpoint differences in
Table~\ref{tab:geometry_ablation}, but it shows that the Euclidean ablation
constitutes a substantial change in the geometry rather than a small
approximation to the Native metric.

\begin{table}[t]
\centering
\caption{
\textbf{Anisotropy of the Adam-induced geometry.}
The full condition number $\kappa(D_t)$ measures the total coordinate-wise
spread, while $p_{95}(D_t)/p_{05}(D_t)$ provides a robust measure less
sensitive to extreme coordinates.
}
\label{tab:adam_anisotropy}
\small
\begin{tabular}{lrrr}
\toprule
PDE
& Step
& $\kappa(D_t)$
& $p_{95}(D_t)/p_{05}(D_t)$ \\
\midrule
Allen--Cahn
& 5k
& $8.21\times10^{6}$
& 21.9 \\

Allen--Cahn
& 10k
& $1.94\times10^{7}$
& 33.8 \\

Wave
& 5k
& $1.13\times10^{6}$
& 16.7 \\

Wave
& 10k
& $1.39\times10^{6}$
& 21.3 \\
\bottomrule
\end{tabular}
\end{table}

The same geometry enters ORACLE twice: it determines the alignment direction
through $P_t^{-1}$ and determines the scale of admissible alignment through
the $P_t$-norm. We therefore next examine how often the native-relative
authority bound becomes active.

\subsection{Geometry and Alignment Authority}
\label{app:authority_regime}

ORACLE bounds the aligned adjustment relative to the Native movement in the
same optimizer-induced metric,
\begin{equation}
    \lVert\widetilde d_t\rVert_{P_t}
    \le
    \beta_t
    \lVert s_t\rVert_{P_t},
    \qquad
    \lVert v\rVert_{P_t}^2
    =
    v^\top P_tv.
    \label{eq:appendix_authority_bound}
\end{equation}

Table~\ref{tab:authority_binding} reports how often this authority boundary is
active among accepted alignments. For full ORACLE, the boundary binds on
$100.0\%$ of accepted Allen--Cahn alignments and $99.2\%$ on Wave. The
metric-aware First-order variant behaves similarly, with rates of $99.7\%$
and $94.8\%$. Under the $P_t=I$ ablation, the corresponding rates fall to
$37.6\%$ and $2.5\%$.

The authority mechanism therefore depends materially on the geometry in which
the Native and aligned movements are measured. Changing $P_t$ affects not only
the direction produced by the alignment objective but also whether a proposed
adjustment reaches the native-relative authority boundary.

The Unbounded variant provides the complementary case: its authority boundary
is absent by construction. Its strong Allen--Cahn endpoint shows that greater
alignment authority can help in some regimes, whereas its smaller improvement
on Wave shows that unrestricted authority is not uniformly preferable across
problems. We therefore treat the authority bound as a mechanism for regulating
alignment magnitude rather than claim that it independently improves every
endpoint.

\begin{table}[t]
\centering
\caption{
\textbf{Activation of the native-relative authority boundary.}
Percentage of accepted alignments for which
Eq.~\eqref{eq:appendix_authority_bound} is active in the matched mechanism
study. Native constructs no alignment candidate, while Unbounded removes the
authority boundary by definition.
}
\label{tab:authority_binding}
\small
\begin{tabular}{lrr}
\toprule
Method
& Allen--Cahn
& Wave \\
\midrule
Native      & --      & --    \\
$P_t=I$     & 37.6\%  & 2.5\% \\
First-order & 99.7\%  & 94.8\% \\
Unbounded   & 0.0\%   & 0.0\% \\
ORACLE      & 100.0\% & 99.2\% \\
\bottomrule
\end{tabular}
\end{table}

\subsection{Held-Out Validation and Commitment}
\label{app:heldout_commitment}

The authority bound controls the magnitude of an alignment candidate but does
not establish that its locally predicted improvement persists on unseen
constraint samples. ORACLE therefore separates candidate construction from
commitment by evaluating the fixed candidate on an independent support
$S'_t$.

We count candidates rejected at this stage as
\begin{equation}
    N_{\mathrm{audit}}
    =
    \sum_{t=1}^{T}
    \mathbf{1}
    \left[
        \mathrm{Audit}_t
        =
        \mathrm{reject}
    \right].
    \label{eq:appendix_audit_count}
\end{equation}

Table~\ref{tab:audit_rejections} reports the resulting counts over the matched
$5$k-step mechanism study. On Allen--Cahn, the $P_t=I$ variant produces
$875$ held-out rejections, compared with $19$ for First-order, $19$ for
Unbounded, and only $4$ for ORACLE. On Wave, the corresponding counts are
$11$, $6$, $1$, and $0$.

These counts provide information not captured by endpoint rel-L2. In
particular, the Euclidean ablation not only ends with substantially larger
solution error but also constructs many more candidates whose local
improvement does not persist on independent support. This diagnostic does not
establish that audit rejection itself causes the endpoint difference; rather,
it shows that the geometry used to construct alignment is associated with
substantially different off-support behavior.

\begin{table}[t]
\centering
\caption{
\textbf{Held-out validation rejections in the matched mechanism study.}
Number of fixed alignment candidates rejected on independent support over
$5$k active optimization steps. Native constructs no alignment candidate and
therefore has no audit decision.
}
\label{tab:audit_rejections}
\small
\begin{tabular}{lrr}
\toprule
Method
& Allen--Cahn
& Wave \\
\midrule
Native      & --  & -- \\
$P_t=I$     & 875 & 11 \\
First-order & 19  & 6  \\
Unbounded   & 19  & 1  \\
ORACLE      & 4   & 0  \\
\bottomrule
\end{tabular}
\end{table}

\subsection{Mechanism Summary}
\label{app:mechanism_summary}

The extended diagnostics support a sequential interpretation of ORACLE.
The Native optimizer first determines the metric in which alignment is
constructed. The same metric defines the scale of the aligned adjustment
relative to the Native proposal. Finally, held-out validation determines
whether the fixed candidate is committed:
\[
\boxed{
P_t
\;\longrightarrow\;
\text{alignment}
\;\longrightarrow\;
\text{authority}
\;\longrightarrow\;
\text{held-out validation}
\;\longrightarrow\;
\text{commitment}.
}
\]

This sequence clarifies the roles of the individual mechanisms.
Replacing $P_t$ by the identity changes both alignment behavior and authority
activation; removing the authority bound changes the allowable intervention
scale; and independent validation can reject candidates that appear locally
acceptable on the construction support.

We therefore do not interpret every component as independently necessary for
every problem. Instead, the complete ORACLE design assigns distinct roles to
the three mechanisms: optimizer geometry determines \emph{how} alignment is
measured, the authority bound limits \emph{how far} the Native movement may be
redirected, and held-out validation determines \emph{whether} the aligned
candidate is ultimately committed.

\subsection{Optimizer-Level Gate Statistics}
\label{app:gate_statistics}

Table~\ref{tab:optimizer_gate_stats} reports how frequently each stage of
ORACLE is reached across the eight Q1 PDE configurations for each optimizer.
\emph{Certified} denotes steps for which the recovered optimizer geometry
passes certification; \emph{Trigger} denotes steps satisfying
$c_t^\top s_t>0$; \emph{Proposal} denotes steps for which a valid alignment
candidate is constructed; and \emph{Accept} denotes steps for which the
candidate passes all validation gates and is committed.

\begin{table}[t]
    \centering
    \caption{\textbf{ORACLE gate statistics across optimizer geometries.}
    Rates are aggregated across the eight PDE configurations associated with
    each optimizer. Certification checks whether the recovered geometry
    faithfully represents the realized native update; trigger marks locally
    adverse native steps; proposal records successfully constructed alignment
    candidates; and accept records corrections ultimately committed after
    validation and held-out audit.}
    \label{tab:optimizer_gate_stats}
    \small
    \setlength{\tabcolsep}{6pt}
    \begin{tabular}{lccccc}
        \toprule
        \textbf{Optimizer} &
        \textbf{Cells} &
        \textbf{Certified} &
        \textbf{Trigger} &
        \textbf{Proposal} &
        \textbf{Accept} \\
        \midrule
        SGD        & 8 & 100.00\% & 59.56\% & 59.55\% & 44.78\% \\
        Adam       & 8 & 100.00\% & 88.46\% & 88.46\% & 84.40\% \\
        SOAP       & 8 & 100.00\% & 94.73\% & 94.73\% & 90.46\% \\
        PSGD--Kron & 8 & 100.00\% & 90.24\% & 90.24\% & 87.24\% \\
        \bottomrule
    \end{tabular}
\end{table}

Geometry certification succeeds throughout these evaluated configurations,
showing that ORACLE can recover the realized optimizer geometry reliably for
all four optimizer families. The subsequent behavior is optimizer dependent:
the alignment trigger fires on $59.56\%$ of SGD steps but on
$88.46$--$94.73\%$ of Adam, SOAP, and PSGD--Kron steps. Once triggered, a
valid alignment proposal is constructed almost always, while the lower final
acceptance rates reflect the nonlinear-agreement and held-out validation gates.
Thus, ORACLE's selectivity arises primarily from detecting whether alignment
is needed and from validating the resulting candidate, rather than from
failures to construct the optimizer-relative correction.

\end{document}